\documentclass{article} 
\usepackage[table,HTML,xcdraw]{xcolor}
\usepackage{iclr2025_conference,times}

\usepackage{hyperref}
\usepackage{url}

\usepackage{multirow}
\usepackage{graphicx} 
\usepackage{array}
\usepackage{booktabs}
\usepackage{algorithm}
\usepackage{algpseudocode}
\usepackage{arydshln}
\usepackage[utf8]{inputenc}
\usepackage[T1]{fontenc}
\usepackage{tikz}
\usepackage{textcomp} 
\usepackage{amssymb,mathrsfs,amsmath}
\usepackage{MnSymbol,bbding,pifont}
\usepackage{graphicx}
\usepackage{subfig}
\usepackage{tcolorbox}
\usepackage{makecell}  
\usepackage{cancel}  
\usepackage{fontawesome}
\usepackage{cleveref}
\usepackage{enumitem}
\crefformat{section}{\S#2#1#3}
\crefformat{subsection}{\S#2#1#3}
\crefformat{subsubsection}{\S#2#1#3}
\crefrangeformat{section}{\S\S#3#1#4 to~#5#2#6}
\crefmultiformat{section}{\S\S#2#1#3}{ and~#2#1#3}{, #2#1#3}{ and~#2#1#3}
\Crefformat{figure}{#2Figure~#1#3}
\Crefmultiformat{figure}{Figs.~#2#1#3}{ and~#2#1#3}{, #2#1#3}{ and~#2#1#3}
\Crefformat{table}{#2Table~#1#3}
\Crefmultiformat{table}{Tabs.~#2#1#3}{ and~#2#1#3}{, #2#1#3}{ and~#2#1#3}
\Crefformat{appendix}{Appx.~\S#2#1#3}
\crefformat{algorithm}{Alg.~#2#1#3}
\Crefformat{equation}{Eq.~(#2#1#3)}
\Crefmultiformat{equation}{Eqs.~(#2#1#3)}{ and~(#2#1#3)}{, (#2#1#3)}{ and~(#2#1#3)}
\usepackage[edges]{forest}
\usepackage[framemethod=tikz]{mdframed}
\usetikzlibrary{arrows.meta,shapes,positioning,shadows,trees}
\usepackage{adjustbox}
\usepackage{tabularx}
\usepackage{textcomp}
\usepackage{array} 

\usepackage{marvosym}
\usepackage{CJKutf8}
\usepackage{fontawesome}

\usepackage{pifont} 
\newcommand{\cmark}{\ding{51}}
\newcommand{\xmark}{\ding{55}}
\definecolor{rliableblue}{HTML}{77AADD}

\title{Reinforcement Learning Foundations for Deep Research Systems: A Survey}

\author{Wenjun Li, Zhi Chen, Jingru Lin, Hannan Cao, Wei Han, Sheng Liang, Zhi Zhang,\\
\textbf{Kuicai Dong, Dexun Li, Chen Zhang, Yong Liu
\vspace{2mm}} \\
Huawei Technologies Co., Ltd
}

\iclrfinalcopy 
\begin{document}

\maketitle

\begin{abstract}
Deep research systems, agentic AI that solve complex, multi-step tasks by coordinating reasoning, search across the open web and user files, and tool use, are moving toward hierarchical deployments with a Planner, Coordinator, and Executors. In practice, training entire stacks end-to-end remains impractical, so most work trains a single planner connected to core tools such as search, browsing, and code. While SFT imparts protocol fidelity, it suffers from imitation and exposure biases and underuses environment feedback. Preference alignment methods such as DPO are schema and proxy-dependent, off-policy, and weak for long-horizon credit assignment and multi-objective trade-offs. A further limitation of SFT and DPO is their reliance on human defined decision points and subskills through schema design and labeled comparisons. Reinforcement learning aligns with closed-loop, tool-interaction research by optimizing trajectory-level policies, enabling exploration, recovery behaviors, and principled credit assignment, and it reduces dependence on such human priors and rater biases.

This survey is, to our knowledge, the first dedicated to the RL foundations of deep research systems. It systematizes recent work along three axes: (i) data synthesis and curation; (ii) RL methods for agentic research covering stability, sample efficiency, long context handling, reward and credit design, multi-objective optimization, and multimodal integration; and (iii) agentic RL training systems and frameworks. We also cover agent architecture and coordination, as well as evaluation and benchmarks, including recent QA, VQA, long-form synthesis, and domain-grounded, tool-interaction tasks. We distill recurring patterns, surface infrastructure bottlenecks, and offer practical guidance for training robust, transparent deep research agents with RL.\\
A curated paper list is available at \href{https://github.com/wenjunli-0/deepresearch-survey}{github.com/wenjunli-0/deepresearch-survey}.
\end{abstract}

\begin{center}
    \vspace{-2mm}
    \includegraphics[width=0.6\textwidth]{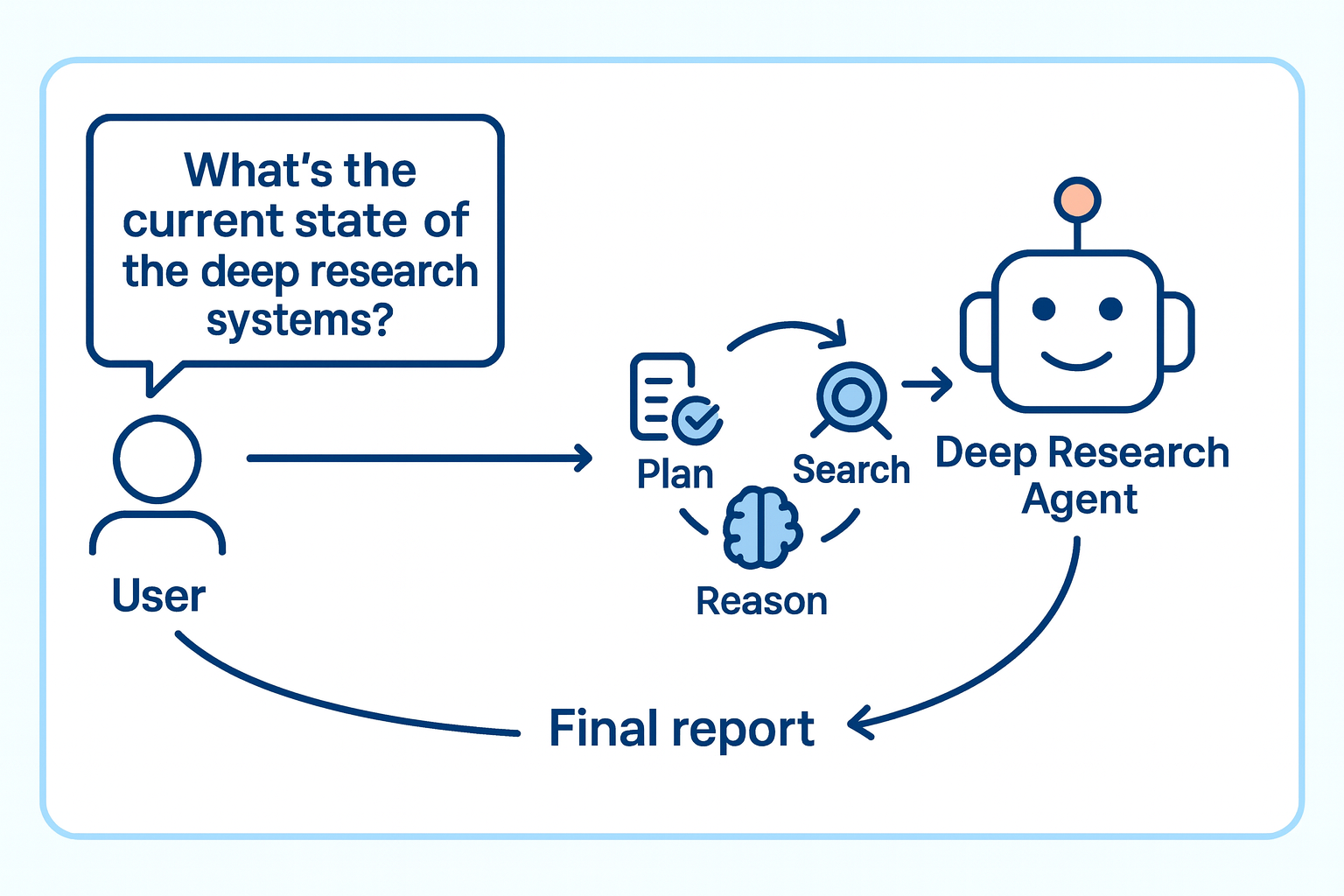} 
    \vspace{2mm}
\end{center}

\clearpage

\pagestyle{empty} 
\setcounter{tocdepth}{3} 
\tableofcontents

\newpage
\pagestyle{plain}
\addtolength{\topmargin}{-25.74332pt} 
\setcounter{page}{1}

\section{Introduction} \label{ch1:intro}
The rapid emergence of deep research systems (e.g., ~\cite{openai2025deepresearch,google2025deepresearch,perplexity2025deepresearch,tongyideepresearchteam2025}), large language models (LLMs) capable of tackling complex, multi-step information-seeking tasks, marks a significant shift in how AI approaches reasoning, execution, and synthesis. In this survey, we focus on information-seeking use cases, as most existing research and products center on this application. We define deep research systems as AI researchers that autonomously plan and carry out multi-step investigations across the open web and user-provided files, iteratively searching, reading, and reasoning as new evidence appears, and ultimately producing either a concise answer for an objective question or a well-structured, citation-backed report for a subjective open question.

\begin{figure}[ht]
  \centering
  \includegraphics[width=0.8\linewidth]{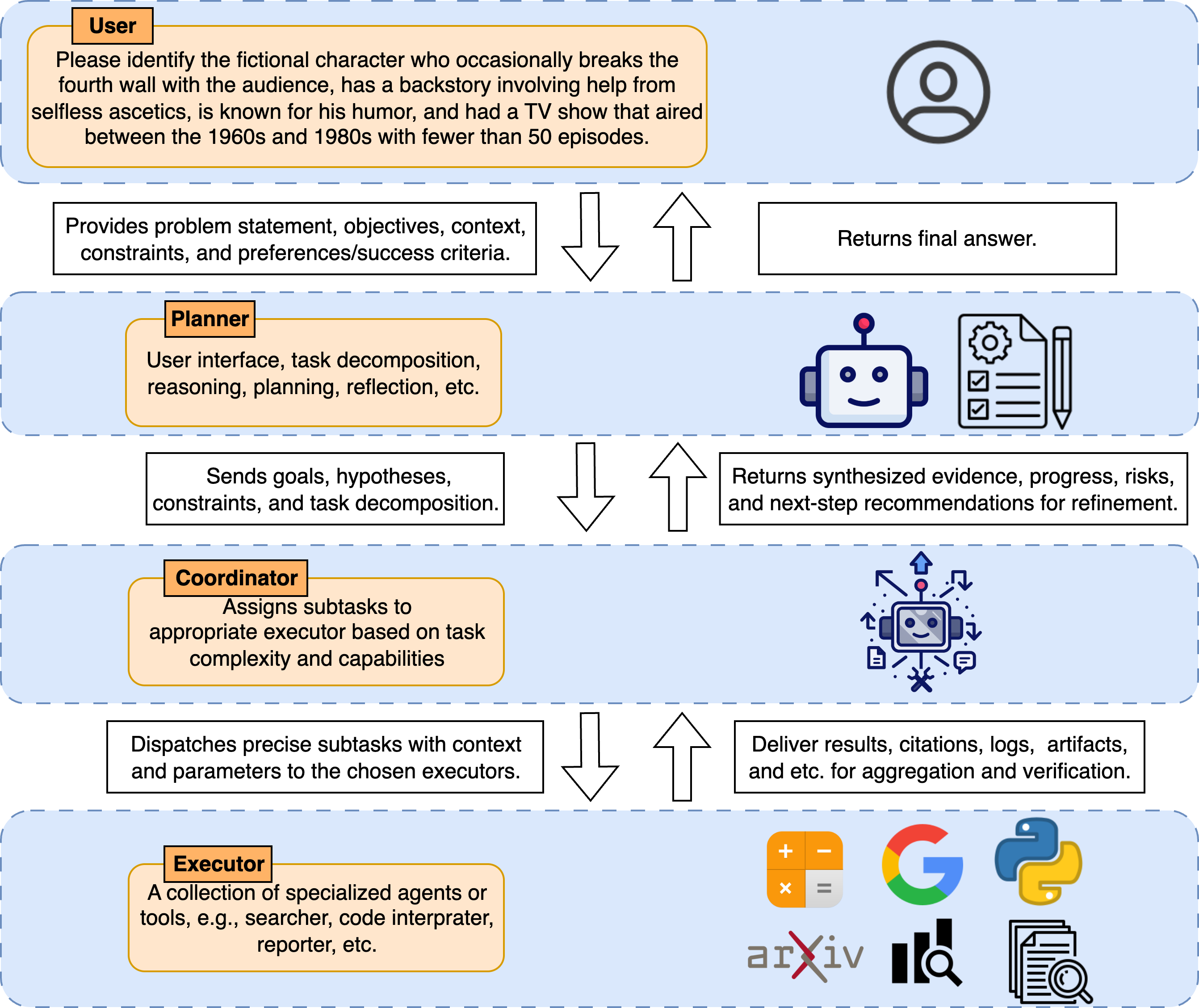}
  \caption{Illustration of the hierarchical deep research system architecture.}
  \label{fig:dr_architecture}
\end{figure}

A trend in both academia~\citep{webthinker_2025,hira_2025,wan2025rema} and industry~\citep{bytedance_deerflow_repo_2025,langchain_open_deep_research_repo_2025,miromind_miroflow_repo_2025} is to move from monolithic agents to hierarchical agent architectures for deep research. Figure~\ref{fig:dr_architecture} mirrors this architecture: a Planner performs step-by-step decomposition and reflection; a Coordinator handles assignment, delegation, aggregation, and verification; and a pool of Executors (i.e., specialized agents and tools) executes grounded actions over the web and files. This separation of concerns decouples strategic planning from execution details, enabling parallelization, plug-and-play expertise (e.g., swapping in better searchers or code runners and scaling to additional tools), and tighter instrumentation for process logging, credit assignment, and auditability. It also keeps the Planner’s global state clean and coherent over long horizons while the Coordinator and Executors handle delegation and grounded actions.

While the hierarchical architecture is attractive for deployment, it is not yet practical to train the entire workflow end-to-end. As a result, most research targets a single model (typically the Planner) wired directly to a small set of fundamental tools (search/browse/code), which collapses rollout length and variance, fits existing RL/SFT/DPO infrastructure, and yields cleaner signals. The training objective is to strengthen long-horizon capabilities in one place (i.e., reasoning, decomposition, tool use, reflection, and synthesis) in an end-to-end manner so that the resulting Planner can later slot into the full hierarchy as a stronger ``brain,'' while coordination and execution remain modular and swappable. Hence, in this survey, we primarily focus on the training of the planner model and will cover the agent architecture and coordination design in Section~\ref{ch:architecture}.

Supervised fine-tuning (SFT; ~\cite{ouyang2022instructgpt,wei2022finetuned}) is an effective way to initialize deep research agents: it is stable, data-efficient, and good at teaching protocol fidelity (e.g., tool-call schemas, response formats), and basic stepwise reasoning patterns. Because SFT optimizes against gold $(x, y)$ pairs, it excels at imparting local behaviors like query rewriting templates, citation style, argument wrapping, and reduces variance early on. The same properties, however, limit performance on multi-turn research tasks. Reference traces are long, composite, and human-authored; imitation induces imitation bias (copying a particular decomposition) and exposure bias (teacher-forced steps hide compounding errors at inference). SFT also leaves environment feedback underused: it cannot directly learn from tool failures, stochastic retrieval, or non-stationary states (e.g., prices, availability). In short, SFT is valuable scaffolding for competencies and interfaces, but not a vehicle for optimizing end-to-end decision quality.

Preference-based methods (e.g., DPO; ~\cite{Rafailov2023DirectPO}) can be pushed beyond single-turn outputs by decomposing agent workflows into labeled steps (e.g., query generation, retrieval selection, synthesis) and learning local preferences at each stage. However, although there are several research works exploring training deep research agents via DPO-based methodologies~\citep{zhang2025reasonrag,zhao2025prefrag,asai2023selfrag}, we think several structural mismatches remain in such approahces. First, DPO optimizes textual alternatives rather than state–action returns: pairwise losses are applied to strings conditioned on prior text, without explicit grounding in environment state (tool results, cache, budgets) or action semantics. This makes credit assignment inherently myopic—it judges which snippet is preferable at that step but cannot attribute downstream success/failure to earlier retrieval or tool-use decisions, nor can it trade off depth of search against cost/latency under partial observability. Second, stepwise DPO inherits schema and proxy dependence: one must hand-design the process decomposition and generate preferences (often with heuristics), which introduces label noise, and brittleness when the unseen task requires a different decomposition. Third, DPO is largely off-policy and offline: it improves on fixed comparisons but does not explore the closed-loop space of actions and tool outcomes, so it struggles to learn recovery behaviors (e.g., when a query returns junk, a site blocks access, or prices shift) and to adapt to non-stationary environments. Finally, multi-objective desiderata (accuracy, calibration, cost, safety) enter only implicitly through rater preferences; DPO provides no principled mechanism to aggregate rewards over long horizons. 

Given the limitations of SFT/DPO-based approaches, both the industry and academia regard reinforcement learning (RL) as a promising pathway towards training deep research agents in an end-to-end manner. Deep research ultimately demands trajectory-level learning in a closed-loop, tool-rich environment: deciding how to decompose problems, when and how to invoke tools, which evidence to trust, when to stop, and how to trade off accuracy, cost, and latency as the state evolves. RL treats the system as a policy over states and actions, enabling end-to-end improvement from environment signals, credit assignment across multi-step traces, and exploration of alternative strategies for search, tool orchestration, recovery, and synthesis. 

Motivated by the shift toward RL for training deep research agents, and the rapid acceleration of progress in this area, we present, to our knowledge, the first survey dedicated to the RL foundations of deep research systems. This survey aims to help researchers systematically understand recent developments, underlying challenges, and key methodological advances. We organize the literature along three primary axes:
\begin{itemize}
\item \textbf{Data Synthesis \& Curation:} Methods for creating and curating complex, high-quality training data, often through synthetic generation, to support multi-step reasoning and tool usage.
\item   n \textbf{RL Methods for Agentic Research:} Works that (i) extend baseline pipelines (e.g., DeepSeek-R1-style~\cite{Deepseek_R1}) to improve stability, sample efficiency, long-context handling; (ii) design rewards and credit assignment that propagate credit across multi-step traces (outcome- vs.\ step-level, composite judges, return decomposition); and (iii) integrate multimodality via backbone VLMs that run iterative perception–reasoning cycles.
\item \textbf{Agentic RL Training Frameworks}: Training deep research agents that interact with tools over long horizons requires a scalable, stable training system. We survey recent open source infrastructures to surface bottlenecks, distill recurring design patterns, and offer practical guidance for composing scalable and reproducible training stacks.
\end{itemize}

Beyond RL training foundations, we highlight two cross-cutting areas that are strategically important:
\begin{itemize}
\item \textbf{Agent Architecture \& Coordination:} Hierarchical, modular, and multi-agent designs that enhance compositional reasoning and division of labor.
\item \textbf{Evaluations \& Benchmarks:} Datasets for assessing deep research systems in holistic, task-rich, tool-interactive settings.
\end{itemize}

Together, these axes provide a cohesive view of the RL-enhanced deep research ecosystem. By tracing advances along each axis, the survey offers a conceptual roadmap for newcomers and a technical reference for researchers aiming to push agentic AI toward robust, real-world problem solving. Figure~\ref{fig:paper-taxonomy} presents the taxonomy and the key papers we survey.
\tikzstyle{defaultbox}=[
rectangle,
draw=black,
rounded corners,
text opacity=1,
minimum height=1.5em,
minimum width=5em,
inner sep=2pt,
align=center,
fill opacity=.5,
fill=blue!10, 
]
\tikzstyle{level1}=[
defaultbox, 
minimum height=1.5em,
text=black,
align=center,
inner xsep=5pt,
inner ysep=4pt,
text width=9em,
anchor=center
]
\tikzstyle{level2}=[
level1,
text width=17em,
]
\tikzstyle{leaf}=[
level2, 
font=\normalsize,
align=left,
text width=42em,
]

\definecolor{Architecture1}{HTML}{a0c4ff}
\definecolor{Architecture2}{HTML}{c0d8ff}
\definecolor{Architecture3}{HTML}{e0f0ff}
\definecolor{RL1}{HTML}{90e0ef}
\definecolor{RL2}{HTML}{a8edf5}
\definecolor{RL3}{HTML}{d0fbff}
\definecolor{Eval1}{HTML}{bdb2ff}
\definecolor{Eval2}{HTML}{d8ccff}
\definecolor{Eval3}{HTML}{f0e8ff}
\definecolor{theme}{HTML}{E0D8D3}

\begin{figure*}[ht]
\centering
\resizebox{\textwidth}{!}
{
    \begin{forest}
        forked edges,
        for tree={
            grow=east,
            reversed=true,
            anchor=base west,
            parent anchor=east,
            child anchor=west,
            base=left,
            font=\large,
            rectangle,
            draw=black,
            rounded corners,
            minimum width=4em,
            edge+={darkgray, line width=1pt},
            s sep=3pt,
            ver/.style={rotate=90, child anchor=north, parent anchor=south, anchor=center, align=center, fill=theme!20,} %
        },
        [RL Foundations{,} Agent Architecture{,} and Evaluations of Deep Research Systems, ver
            [RL Foundations, level1, fill=RL1, minimum width=12em
                [Data Synthesis \& Curation~(\S\ref{ch2:data}), level2, fill=RL2
                    [
                        DeepResearcher~\citep{deepresearcher_2025}{,} 
                        R1-Searcher~\citep{song2025r1}{,} 
                        WebPuzzle~\citep{shi2025pangu}{,} 
                        R-Search~\citep{shi2025reinforcement}{,} 
                        SearchExpert~\citep{li2025enhancing}{,} 
                        Go-Browse~\citep{gandhi2025go}{,} 
                        StepSearch~\citep{wang2025stepsearch}{,} 
                        MEM1~\citep{zhou2025mem1}{,} 
                        SWiRL~\citep{goldie2025synthetic}{,} 
                        WebDancer~\citep{wu2025webdancer}{,} 
                        WebSailor~\citep{li2025websailor}{,}
                        WebShaper~\citep{tao2025webshaper}{,}
                        WebWatcher~\citep{geng2025webwatcher}
                        , leaf, fill=RL3]
                ]
                [Training Regime and Optimization Structure~(\S\ref{ch3.1:regime}), level2, fill=RL2
                    [
                        Search-R1~\citep{jin2025search}{,} 
                        ReSearch~\citep{chen2025researchlearningreasonsearch}{,}
                        R1-Searcher~\citep{song2025r1}{,}
                        WebSailor~\citep{li2025websailor}{,}
                        DeepDiver~\citep{shi2025pangu}{,}
                        ZeroSearch~\citep{sun2025zerosearch}{,}
                        MEM1~\citep{zhou2025mem1}{,}
                        RAG-R1~\citep{tan2025ragr1}{,}
                        Reasoning-Table~\citep{lei2025reasoningtable}{,}
                        FrugalRAG~\citep{java2025frugalrag}{,}
                        EvolveSearch~\citep{zhang2025evolvesearch}{,}
                        ARPO~\citep{dong2025ARPO}{,}
                        Writing-RL~\citep{lei2025writing}
                        , leaf, fill=RL3]
                ]
                [Reward Design and Credit Assignment~(\S\ref{ch3.2:reward}), level2, fill=RL2
                    [
                        s3~\citep{jiang2025s3}{,}
                        AutoRefine~\citep{shi2025searchrefinethinkautonomous}{,}
                        MT-GRPO~\citep{zeng2025MTGRPO}{,}
                        IKEA~\citep{huang2025IKEA}{,}
                        ARTIST~\citep{singh2025artist}{,}
                        R1-Searcher++~\citep{song2025r1searcher++}{,}
                        StepSearch~\citep{wang2025stepsearch}{,}
                        $O^{2}$-Searcher~\citep{mei2025o2searcher}{,} 
                        R-Search~\citep{zhao2025rsearchempoweringllmreasoning}
                        , leaf, fill=RL3]
                ]
                [Multimodal Research Agents~(\S\ref{ch3.3:multimodal}), level2, fill=RL2
                    [
                        VRAG-RL~\citep{wang2025vragrl}{,} 
                        Visual-ARFT~\citep{liu2025visual}{,}
                        WebWatcher~\citep{geng2025webwatcher}{,}
                        MMSearch-R1~\citep{wu2025mmsearch}{,}
                        V-ToolRL~\citep{su2025vtoolrl}{,}
                        VTool-R1~\citep{wu2025vtool}
                        , leaf, fill=RL3]
                ]
                [Agentic RL Training Frameworks:~(\S\ref{ch4:infra}), level2, fill=RL2
                    [
                        Agent Lightning~\citep{luo2025agentlightning}{,} 
                        AREAL~\citep{fu2025areal}{,} 
                        AWorld~\citep{yu2025aworld}{,} 
                        OpenR~\citep{wang2024openr}{,} 
                        rLLM~\citep{rllm2025}{,} 
                        ROLL~\citep{wang2025roll}{,} 
                        SLIME~\citep{2025SLIME}{,} 
                        Verifiers~\citep{2025verifiers}{,}
                        verl~\citep{sheng2025verl}
                        , leaf, fill=RL3]
                ]
            ]   
            [Agent Architecture and Coordination~(\S\ref{ch:architecture}), level1, fill=Architecture1, minimum width=12em
                [Open-Source Architectures, level2,fill=Architecture2
                    [
                        Aomni Open Deep Research~\citep{dzhng_deep_research_repo_2025}{,} 
                        ByteDance DeerFlow~\citep{bytedance_deerflow_repo_2025}{,} 
                        LangChain Open Deep Research~\citep{langchain_open_deep_research_repo_2025}{,} 
                        MiroMindAI MiroFlow~\citep{miromind_miroflow_repo_2025}
                    , leaf,fill=Architecture3]
                ]
                [Academic Architectures, level2,fill=Architecture2
                    [
                        OWL~\citep{owl_2025}{,} 
                        CoA~\citep{li2025chainofagents}{,} 
                        PaSa~\citep{he-etal-2025-pasa}{,} 
                        WebThinker~\citep{webthinker_2025}{,} 
                        HiRA~\citep{hira_2025}{,} 
                        DeepResearcher~\citep{deepresearcher_2025}
                    , leaf,fill=Architecture3]
                ]
                [RL for Multi-Agent Coordination, level2,fill=Architecture2
                    [
                        MHGPO~\citep{chen2025mhgpo}{,} 
                        MMOA-RAG~\citep{chen2025mmoa}{,} 
                        Optimas~\citep{wu2025optimas}
                    , leaf,fill=Architecture3]
                ]
            ] 
            [Evaluation and Benchmarks~(\S\ref{ch:eval}), level1,fill=Eval1, minimum width=12em
                [QA and VQA Benchmarks, level2,fill=Eval2
                    [
                        HotpotQA~\citep{yang2018hotpotqa}{,}
                        2Wiki~\citep{xanh2020_2wikimultihop}{,}
                        Natural Questions~\citep{kwiatkowski2019natural}{,} 
                        MuSiQue~\citep{trivedi2021musique}{,} 
                        FEVER~\citep{thorne-etal-2018-fever}{,} 
                        Bamboogle~\citep{press2022measuring}{,} 
                        FRAMES~\citep{krishna-etal-2025-fact}{,} 
                        BrowseComp~\citep{wei2025browsecomp}{,} 
                        BrowseComp-ZH~\citep{zhou2025browsecomp}{,}
                        InfoDeepSeek~\citep{xi2025infodeepseek}{,}
                        WebWalker~\citep{wu2025webwalker}{,}
                        WideSearch~\citep{wong2025widesearchbenchmarkingagenticbroad}{,}
                        MMSearch~\citep{jiang2025mmsearch}{,}
                        MMDocIR~\citep{dong2025mmdocir}{,}
                        MRAGMG-Bench~\citep{yu2025mramg}{,}
                        M$^2$RAG~\citep{liu2025benchmarkingretrievalaugmentedgenerationmultimodal}{,}
                        MMDocRAG~\citep{dong2025mmdocrag}{,}
                        MM-BrowseComp~\citep{li2025mmbrowsecompcomprehensivebenchmarkmultimodal}{,}
                        Omni-Bench~\citep{li2024omnibench}
                    , leaf,fill=Eval3]
                ]
                [Long-form Text Benchmarks, level2,fill=Eval2
                    [
                        ProxyQA~\citep{tan-etal-2024-proxyqa}{,}
                        WritingBench~\citep{wu2025writingbench}{,}
                        LongEval~\citep{alkhalifa2024longeval}{,}
                        DeepResearch Bench~\citep{futuresearch2025deepresearchbench}
                    , leaf,fill=Eval3]
                ]
                [Domain-Grounded Benchmarks, level2,fill=Eval2
                    [
                        Xbench~\citep{chen2025xbench}{,}
                        $\tau^2$-Bench~\citep{barres2025tau2benchevaluatingconversationalagents}{,}
                        Finance Agent Benchmark~\citep{bigeard2025financeagentbenchmarkbenchmarking}{,}
                        FinGAIA~\citep{zeng2025fingaiachinesebenchmarkai}{,}
                        OdysseyBench~\citep{wang2025odysseybenchevaluatingllmagents}
                    , leaf,fill=Eval3]
                ]
            ] 
        ]
    \end{forest}
}
\caption{The organizational structure of the survey and representative papers under each branch.}
\label{fig:paper-taxonomy}
\end{figure*}
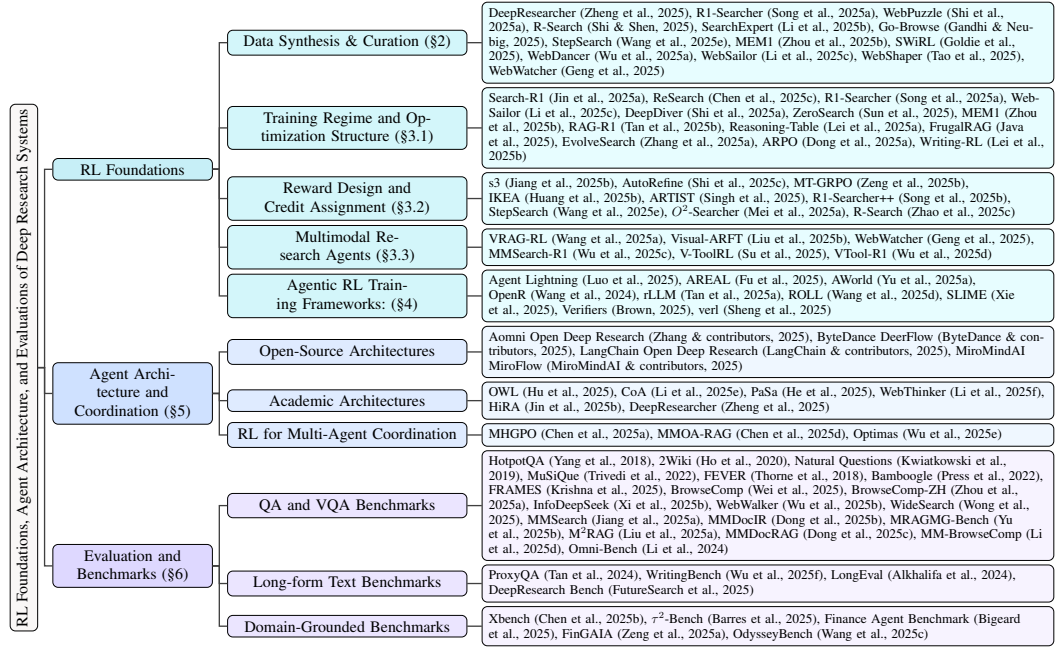

\paragraph{Positioning and contributions} 
Unlike concurrent surveys \citep{huang2025surveydeep,li2025surveyaisearch,xi2025survey,xu202surveycomprehensive,li2025surveytowards,zhang2025surveydeep,zhang2025surveylandscape}, we adopt a training-first, RL-centric perspective. We (i) explain why SFT/DPO misalign with closed-loop, tool-interactive research and motivate end-to-end RL at the planner; (ii) present, to our knowledge, the first taxonomy dedicated to RL foundations for deep research, spanning data synthesis \& curation, RL methods for agentic research, and training frameworks; (iii) treat agentic RL training as a systems problem, surfacing infrastructure bottlenecks and scalable training; (iv) connect training to deployment via a planner-centric training vs. hierarchical execution decoupling, and provide a comprehensive, in-depth synthesis of evaluations and benchmarks. Compared with contemporaneous overviews, our survey is more narrowly focused on RL and delivers deeper data, algorithmic, and infrastructure insights. Together, our survey offers a unified blueprint and actionable guidance for training robust deep research agents via RL.

\paragraph{Timeframe and inclusion criteria}
We survey RL-based training for deep research agents released after February 2025 (post–DeepSeek R1) through September 2025 (manuscript cut-off), covering RL training pillars and agent architecture and coordination designs. For benchmarks papers, we cite canonical QA/VQA and long-form text benchmarks developed in recent years, while domain-grounded, tool-interactive benchmarks are restricted to 2025 vintages.

\section{Data Synthesis \& Curation} \label{ch2:data}
Data synthesis has become an increasingly critical component of data-driven training paradigms, particularly in the development of modern AI systems. With the rise of generative models, synthetic data generation has become more accessible and cost-effective. However, curating high-quality synthetic data for training deep research agents remains a challenging task, especially when aiming to support complex reasoning, tool use, and multi-step decision-making.

In this section, we first discuss how RL training data differs from that of SFT/DPO training, and examine the essential properties that synthetic data should exhibit to effectively support the training of deep research agents. We then provide an overview of recent works that explore various techniques for constructing complex queries and curating training data. We introduce a taxonomy of data complexity to better characterize the difficulty and structure of synthetic tasks. Finally, we outline the insights and open challenges in this rapidly evolving area.

\begin{table*}[ht]
\centering
\small
\caption{Summary of papers in this section. New Dataset(s) indicates whether a named dataset was newly released (as opposed to only curating or augmenting existing data).}
\begin{tabularx}
{\textwidth}{@{}p{6.0cm} >{\raggedright\arraybackslash}X@{}}
\toprule
\textbf{Paper} & \textbf{New Dataset(s)} \\
\midrule
\multicolumn{2}{@{}l}{\textbf{\emph{Construction methods with new datasets}}}\\
\textit{DeepDiver}~\citep{shi2025pangu} & Yes — \textit{WebPuzzle}. \\
\textit{WebDancer}~\citep{wu2025webdancer} & Yes — \textit{CrawlQA}, \textit{E2HQA}. \\
\textit{WebSailor}~\citep{li2025websailor} & Yes — \textit{SailorFog-QA}. \\
\textit{WebShaper}~\citep{tao2025webshaper} & Yes — \textit{WebShaper}. \\
\textit{WebWatcher}~\citep{geng2025webwatcher} & Yes — \textit{BrowseComp-VL}. \\
\textit{InfoSeek}~\citep{xia2025opendatasynthesisdeep} & Yes — \textit{InfoSeek}. \\
\addlinespace[0.6ex]
\multicolumn{2}{@{}l}{\textbf{\emph{Systems / pipelines (no new dataset)}}}\\
\textit{R-Search}~\citep{shi2025reinforcement} & No — uses fresh corpora. \\
\textit{SearchExpert}~\citep{li2025enhancing} & No — constructed tasks. \\
\textit{SWiRL}~\citep{goldie2025synthetic} & No — rollout prefixes. \\
\textit{Go-Browse}~\citep{gandhi2025go} & No — curated existing datasets. \\
\textit{StepSearch}~\citep{wang2025stepsearch} & No — augments \textit{MuSiQue}. \\
\textit{Search-R1}~\citep{jin2025search} & No — uses \textit{NQ} + \textit{HotpotQA}. \\
\textit{R1-Searcher}~\citep{song2025r1} & No — difficulty labels on existing data. \\
\textit{MEM1}~\citep{zhou2025mem1} & No — synthesized from existing data. \\
\textit{DeepResearcher}~\citep{deepresearcher_2025} & No — curated existing datasets. \\
\bottomrule
\end{tabularx}
\label{tab:dr-data-summary}
\end{table*}

\subsection{Format of RL Training Data}
We begin by clarifying the purpose, format, and requirements of RL training data for deep research agents. This background frames the discussion that follows and clarifies the terminology we use throughout.

RL optimizes \emph{goal achievement in an environment}. Formally, the deep research agent learns a policy \(\pi_\theta\) that maximizes expected return under tool-use interactions. The RL training setup comprises \emph{environment, tasks, and reward}. Let the task set be
\[
\mathcal{Q}=\{q_j\}_{j=1}^M,
\]
often multi-step questions/tasks. The interactive environment is
\[
\mathcal{M}=(\mathcal{S},\mathcal{A},\mathcal{O},P,r),
\]
where \(\mathcal{S}\) are states, \(\mathcal{A}\) are actions (e.g., tool calls like search, browse, code interpreter), \(\mathcal{O}\) are observations (pages/snippets/images), \(P\) defines state-transition dynamics, and \(r\) yields rewards. A trajectory
\[
\tau=(o_0,a_0,o_1,a_1,\ldots,o_T,a_T)
\]
consists of observations and actions during task execution. Outcome-only rewards take the form
\[
R(q,\tau)=g\!\left(\hat{y}(\tau),\,y_q^\star\right)
\quad\text{(\(g(\cdot)\) can be e.g., EM/F1),}
\]
and can be augmented with stepwise/process signals \(r_t(q,\tau)\) (e.g., feasibility checks). Crucially, RL does \emph{not} require expert trajectories; \((q, y_q^\star)\) plus a reliable \(g(\cdot)\) suffice.

Most RL pipelines for deep research adopt an DeepSeek-R1-style regime and use a small ``SFT'' set to cold-start the policy (see Section~\ref{ch:rl}). Unlike pure SFT, this data’s role is interface compliance and scaffolded rollouts, not end-to-end imitation. Each example is a full trajectory that (i) plans and decomposes the query (\texttt{$<$think$>$...$<$/think$>$}); (ii) invokes tools with validated argument schemas (\texttt{$<$tool\_name$>$...$<$/tool\_name$>$}); (iii) parses and interprets intermediate outputs (\texttt{$<$result$>$...$<$/result$>$}); and (iv) synthesizes the final response (\texttt{$<$answer$>$...$<$/answer$>$}). The emphasis is on correct calling conventions, step ordering, and smooth transitions between reasoning, tool interaction, and answer synthesis; subsequent training then replaces imitation with end-to-end training from the initial query, through tool-mediated interactions, to final rewards.

RL training data for deep research agents requires: (i) tasks that resist parametric recall, e.g., a cross-document/recency constraint \(\chi(q)\ge 2\) and a contamination predicate \(C(q)=0\) (not answerable from parametric memory alone); (ii) \emph{cheap, stable, objective} rewards \(R\) (e.g., exact answers, checklists, functional tests). For fair comparison, RL pipelines also fix retrieval settings and enforce step/time caps on rollouts.

RL performance hinges on \emph{which tasks} are posed and \emph{how} success is verified and selected. We therefore separate:
\[
\operatorname{Construct}(\mathcal{C}) \;\to\; \tilde{\mathcal{Q}},
\]
which maps corpora/web sources \(\mathcal{C}\) into \emph{candidate tasks} that demand multi-step reasoning and tool use; and
\[
\operatorname{Curate}(\tilde{\mathcal{Q}})=\{\,q\in\tilde{\mathcal{Q}}:\;\bigwedge_{f\in\mathcal{F}} f(q)=1\,\},
\]
a filtering/scheduling pipeline with optional curriculum \(\mu(q)\) implementing contamination/novelty gates, outcome/process verification, and difficulty staging.

\subsection{Constructing Complex Queries and Curating Data}
We analyze data preparation for deep research systems along two complementary axes—how complex queries are constructed and how the resulting data are curated. We first cover construction strategies (what goes into $\tilde{\mathcal{Q}}$), then turn to curation (what remains and how it is scheduled).

\paragraph{Construction Strategies}
To elicit and cultivate long-horizon capabilities—robust reasoning, iterative tool interaction, reflection, and synthesis—we require genuinely complex queries that compel the model to perform multiple rounds of planning, evidence gathering, and verification. To avoid ``shortcut'' solutions (e.g., answers obtainable via a single lookup or memorized fact), a line of work focuses on strategies that deliberately increase task difficulty while preserving verifiability. We group these strategies into three categories, outlined below:
\begin{itemize}
\item[1.] \textbf{Cross-Source Composition (often recency-aware).} These methods author questions that require integrating evidence across multiple sources—typically drawn from recent news/papers/webpages to push beyond the model’s parametric memory. \textit{R-Search} clusters fresh documents (news + arXiv) and trains a single LLM to plan, execute multi-source search, and synthesize an answer in one pass; the learning objective explicitly couples reasoning with structured, stepwise search. \textit{WebPuzzle} generates cross-page inverted QA from multiple open-web pages and further assigns pass@k difficulty tags to winnow easy items, producing a curated subset for RL. \textit{SearchExpert} also starts from fresh crawls but is plan-centric: it produces code-level search DAGs and converts them into compact natural-language DAGs for SFT; harder cases are used for RL with a search-feedback reward tied to retrieved evidence quality.
\item[2.] \textbf{Structure-Driven Path Growth (graph/set navigation).} Here, solution length grows by expanding over hyperlink graphs or by composing formal set operations. \textit{CrawlQA} grows paths by recursive link traversal from authoritative roots (arXiv/GitHub/Wikipedia) to emulate human browsing, then synthesizes typed questions (e.g., COUNT, MULTI-HOP, INTERSECTION) from the visited page set—lengthening solution paths via browsing rather than obfuscation. \textit{WebSailor} builds dense site graphs via random walks from rare entities, then samples subgraphs to create SailorFog-QA—questions that demand multi-hop browsing and non-linear synthesis. \textit{WebWatcher} constructs a Wikipedia hyperlink graph and turns selected items into image-grounded VQA to require cross-modal retrieval. \textit{WebShaper} replaces literal graph traversal with a set-theoretic formalism—Knowledge Projections—and a layer-wise Expander that controls reasoning depth while avoiding shortcut paths. \textit{Go-Browse} treats data collection as structured exploration over a reusable URL graph (NavExplorer, PageExplorer), keeping only tasks that pass a feasibility check judged by a VLM. 
\item[3.] \textbf{Difficulty Staging by Transformation/Evolution.} These create easy-to-hard progressions via rewriting or step-level supervision. \textit{E2HQA} iteratively rewrites a simple seed by replacing entities with constraints mined from the web while preserving the final answer, so each iteration adds hops. \textit{StepSearch} augments MuSiQue with sub-question trajectories and trains with information-gain rewards and redundancy penalties per step, encouraging deeper, better-targeted queries. \textit{MEM1} increases workload by bundling multiple independent questions into a single multi-objective prompt (harder via breadth rather than tighter cross-step dependency). \textit{SWiRL} rolls out multi-step tool-use trajectories with an open-source model, then splits each k-step trace into k prefixes to form an implicit prefix curriculum. 
\end{itemize}

Beyond construction and curation, some work introduces a cross-cutting modifier: \textbf{obfuscation}. \textit{DeepDiver}, \textit{WebSailor}, and \textit{WebWatcher} deliberately mask entities or fuzz attributes (e.g., vague dates, indirect descriptors) so direct lookup fails, forcing multi-step discovery. This is an orthogonal knob to raise difficulty and can be combined with any of the construction and curation choices above. 

\paragraph{Curation Strategies}
Orthogonal to constructing new complex queries, other work designs selection procedures that decide what stays for training. We survey how examples are filtered/selected as follows: 
\begin{itemize}
\item[1.] \textbf{Contamination and Novelty Gates.} These filters remove samples solvable from parametric memory or by a single document so that training focuses on truly complex tasks. \textit{DeepResearcher} drops any question that the base model answers within pass@10, ensuring novelty beyond internal knowledge. \textit{SearchExpert} retains only items that require its crawled context, explicitly rejecting cases answerable without retrieval from the constructed corpus.
\item[2.] \textbf{Outcome-Verified Selection.} This keeps items whose rollouts are either demonstrably solvable or demonstrably nontrivial under a verifiable success test. \textit{Search-R1} optimizes and reports final answer EM on a unified NQ+HotpotQA pool, using the same objective to filter useful rollouts for continued training. \textit{WebPuzzle} tags each item with pass@k difficulty and filters out easy cases to yield a harder RL subset. \textit{Go-Browse} requires at least one verified success judged by a VLM before an item is admitted, which both proves feasibility and guards against degenerate tasks that waste budget.

\item[3.] \textbf{Process-Quality Filtering.} Rather than judging only final answers, these methods validate intermediate steps so credit aligns with sound action and reasoning. \textit{SWiRL} introduces an LLM-as-judge process filter to check coherence and rationale alignment across steps. \textit{StepSearch} augments MuSiQue with sub question trajectories and uses information gain rewards and redundancy penalties at each step, effectively filtering for processes that add evidence rather than repeat it. \textit{Go-Browse} also applies a feasibility judge during exploration, discarding paths that fail stepwise checks even if a final answer could be guessed.

\item[4.] \textbf{Difficulty-Aware Curricula and Scheduling.} After filtering, examples are staged so training time concentrates on the band that moves learning the most. \textit{R1-Searcher} labels items by observed rollout counts on Hotpot and 2Wiki (easy, medium, hard) and upweights medium and hard during training. \textit{WebPuzzle} samples a hard mix after tagging to keep batches informative.
\end{itemize}

\subsection{Classification of Query Complexity}

\begin{figure}[t]
  \centering
  \includegraphics[width=0.8\linewidth]{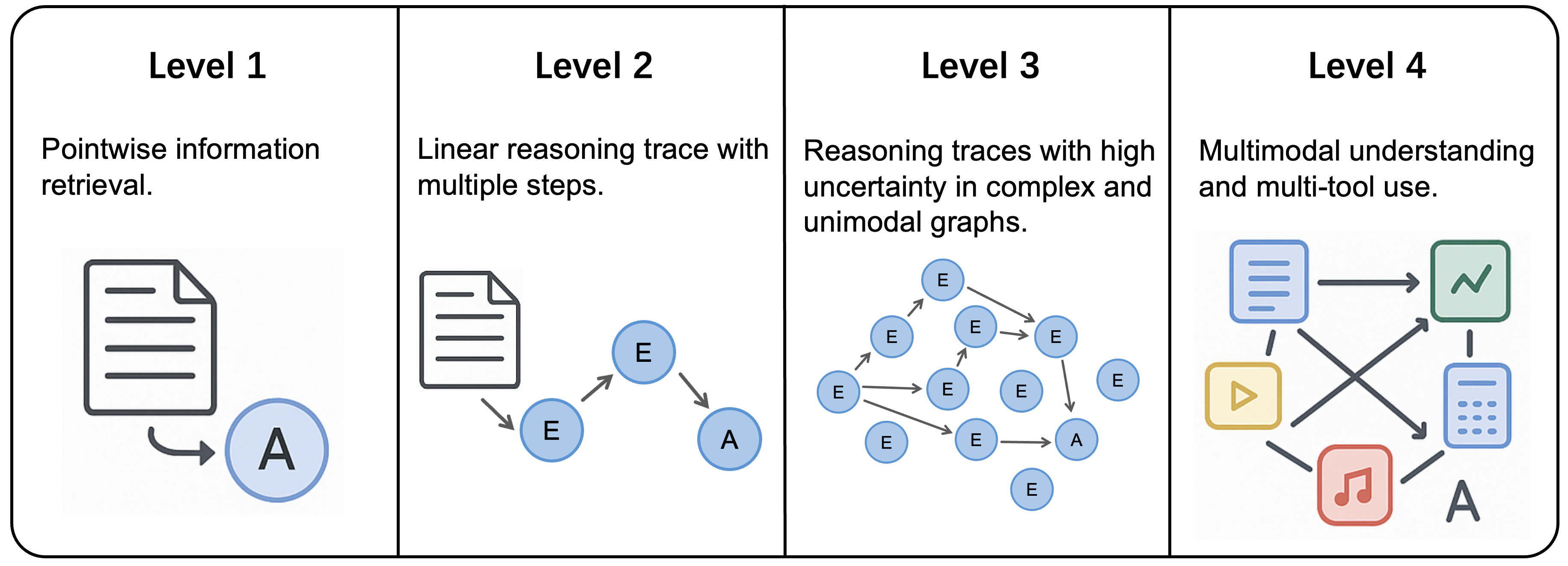}
  \caption{Illustration of QA Task Complexity Levels.}
  \label{fig:data_classification}
\end{figure}

After surveying recent advancements in constructing and curating RL training data for deep research agents, we propose the following classification of QA tasks based on their complexity. This taxonomy provides a common language to (i) stratify datasets for curriculum design, (ii) report results by difficulty band, and (iii) diagnose failure modes in long-horizon training:
\begin{itemize}
\item[Level 1:] Questions with low uncertainty that can be answered directly using the model’s internal knowledge or a single, straightforward web search. Examples include factual queries about natural phenomena or simple weather-checking questions. Example datasets: SimpleQA~\citep{wei2024simpleqa}, TriviaQA~\citep{joshi2017triviaqa}.
\item[Level 2:] Multi-hop questions that require multiple searches but follow a clear and well-defined reasoning path. These questions can be solved through a structured sequence of logical steps. For instance, questions generated via easy-to-hard reverse construction techniques fall into this category. Example datasets: HotpotQA~\citep{yang2018hotpotqa}, Bamboogle~\citep{press2022measuring}.
\item[Level 3:] Questions characterized by both high uncertainty and high difficulty in reducing that uncertainty. The involved entities are connected in complex and often emergent ways, with no pre-defined reasoning path. Solving these tasks requires extensive exploration and rigorous cross-validation. Importantly, these questions remain \emph{unimodal}, meaning that the entire reasoning chain and all necessary clues can be resolved through textual search alone. Example datasets: SailorFog-QA, WebShaper.
\item[Level 4:] Questions that require both \emph{multimodal} understanding and multi-tool orchestration, extending Level 3 in complexity. The input question itself may be presented across different modalities (e.g., text, images, audio), requiring the agent to first interpret the query correctly. Solving the task demands coordinating multiple tools (e.g., image recognition, audio transcription, code execution) and integrating evidence from diverse modalities to produce a final answer. Example datasets: WebWatcher, FactualVQA~\citep{wu2025mmsearch}.
\end{itemize}

\subsection{Discussion}
We suggest using the levels as a practical curriculum: warm-start on Levels 1–2 to establish interface compliance and basic planning, advance to Level 3 for exploration and cross-validation skills, and add Level 4 last for multimodal coordination. Report learning curves and sample efficiency by level to make these stages visible. Treat ``construction'' and ``curation'' as complementary levers observed in the literature: construction increases structural difficulty (hops, branching, recency, modality), while curation improves signal quality (decontamination, process checks, difficulty mixing).

We present the following questions for future research: (i) How can we generate large-scale, complex, multimodal queries at low cost while ensuring feasibility and verifiability? (ii) Active, difficulty-aware task generation during training. Can we close the loop so the agent (and its value/uncertainty estimates) drives on-the-fly construction and curation—selecting recency windows, obfuscation knobs, and curricula to maximize sample efficiency? 

\section{RL Methods for Agentic Research} \label{ch:rl}

\begin{table*}[ht]
  \footnotesize
  \centering
  \caption{Summary of papers in Section~\ref{ch3.1:regime}.}
  \begin{tabularx}{\textwidth}{
  @{}p{2.8cm}@{\hspace{0.2cm}}p{3.8cm}
  @{\hspace{0.3cm}}>{\centering\arraybackslash}p{1.0cm}
  p{3cm}@{\hspace{0.2cm}}p{2.5cm}@{} 
  }
    \toprule
    \textbf{Work} & \textbf{Policy Model} & \textbf{Cold Start?} & \textbf{Reward Type} & \textbf{RL Optimizer}\\
    \midrule
    \textit{Search-R1}\newline\citep{jin2025search}   &Qwen2.5-7B-Base/Instruct      & \xmark & Outcome &PPO,\newline GRPO \\
    
    \addlinespace
    \textit{ReSearch}\newline\citep{chen2025researchlearningreasonsearch}    &Qwen2.5-32B-Base/Instruct      & \xmark & Outcome+Format &GRPO \\
    
    \addlinespace
    \textit{R1-Searcher}\newline\citep{song2025r1} &Qwen2.5-7B-Base,\newline Llama-3.1-8B-Instruct       & \xmark & Outcome+Format &Reinforce++,\newline GRPO \\
    
    \addlinespace
    \textit{WebSailor}\newline\citep{li2025websailor}   &Qwen2.5-72B-Base               & \cmark & Outcome+Format &DUPO\newline (GRPO-variant) \\
    
    \addlinespace
    \textit{DeepDiver}\newline\citep{shi2025pangu}   &Qwen2.5-7B-Instruct,\newline Pangu-7B-Reasoner               & \cmark & Outcome+Format\newline+Retrieval\_Necessity &GRPO \\
    
    \addlinespace
    \textit{ZeroSearch}\newline\citep{sun2025zerosearch}   &Qwen2.5-7B-Base/Instruct,\newline LLama-3.2-3B-Base/Instruct   & \xmark & Outcome &REINFORCE,\newline GRPO, PPO \\

    \addlinespace
    \textit{ASearcher}\newline\citep{gao2025asearcher}   &Qwen2.5-14B-Base,\newline QwQ-32B   & \xmark & Outcome+Format &GRPO \\

    \addlinespace
    \textit{SSRL}\newline\citep{fan2025ssrl}   &Llama-3.1-8B-Base/Instruct,\newline Qwen2.5-7B-Instruct   & \xmark & Outcome+Format &GRPO \\

    \addlinespace
    \textit{MEM1}\newline\citep{zhou2025mem1}   &Qwen2.5-7B-Base   & \xmark & Outcome &PPO \\
    
    \addlinespace
    \textit{RAG-R1}\newline\citep{tan2025ragr1}   &Qwen2.5-7B-Instruct   & \cmark & Outcome &PPO \\
    
    \addlinespace
    \textit{Reasoning-Table}\newline\citep{lei2025reasoningtable}   &Qwen2.5-7B-Instruct,\newline Qwen2.5-Coder   & \cmark & Outcome+Format\newline+Position &GRPO \\
    
    \addlinespace
    \textit{FrugalRAG}\newline\citep{java2025frugalrag}   &Qwen2.5-7B-Instruct,\newline Llama-3.1-8B-Instruct   & \cmark & Outcome\newline+Retrieval\_Necessity &GRPO \\
    
    \addlinespace
    \textit{EvolveSearch}\newline\citep{zhang2025evolvesearch}   &Qwen2.5-7B-Instruct   & \cmark & Outcome+Format &GRPO \\
    
    \addlinespace
    \textit{ARPO}\newline\citep{dong2025ARPO}   &Qwen3-14B-Base,\newline Llama-3.1-8B-Instruct  & \cmark & Outcome &ARPO\newline (GRPO-variant) \\
    
    \addlinespace
    \textit{Writing-RL}\newline\citep{lei2025writing}   &Qwen2.5-7B-finetuned,\newline Llama-3.1-8B-finetuned  & \cmark & Outcome &PPO \\

    \bottomrule
  \end{tabularx}
  \label{tab:rl-regime}
\end{table*}

With high-quality synthetic data in place, the next step is designing effective RL training pipelines for deep research agents. As established in Section~\ref{ch1:intro}, this chapter focuses on recent advances in RL-based training (rather than SFT/DPO), emphasizing end-to-end learning that strengthens long-horizon reasoning, planning, tool use, reflection, and synthesis. 

We organize the literature into three threads: (i) Training Regime and Optimization Structure: works that go beyond the baseline pipeline (i.e., DeepSeek-R1-style~\cite{Deepseek_R1} and Search-R1-style~\cite{jin2025search}) to improve stability, sample efficiency, long-context handling, etc; (ii) Reward Design and Credit Assignment: approaches that determine what behaviors are reinforced and how credit is propagated across multi-step traces, including outcome-level vs. step-level rewards, novel reward designs, and return decomposition; (iii) Multimodal Research Agents: end-to-end multimodal agents that use a multimodal backbone (e.g., vision language models (VLMs)) to perform iterative perception–reasoning cycles; we prioritize works that internalize multimodal competence rather than delegating it primarily to external tools (e.g., pure OCR calls).

For quick reference, Table~\ref{tab:rl-regime},~\ref{tab:rl-reward},~\ref{tab:rl-multimodal} summarize the papers covered in Section~\ref{ch:rl}, reporting their backbone policy models, whether a cold start is used before RL (e.g., SFT/RSFT/none), and the reward types employed (outcome/format/etc). For brevity, we report only the largest backbone size from each model family when multiple scales are used in the same paper (e.g., if a paper trains on both Qwen2.5-3B/7B and Llama-3-8B/70B, we list only Qwen2.5-7B and Llama-3-70B). Cold start refers to the SFT/RSFT to learn reasoning skeletons, tool invocation, and answer formats. Outcome reward evaluates the correctness of the final answer (e.g., EM/F1), while format reward checks compliance with reasoning skeletons and parsing requirements (e.g., final answer and tool invocations formats). Additional reward designs are discussed in Section~\ref{ch3.2:reward}.

\subsection{Training Regime and Optimization Structure} \label{ch3.1:regime}
In this section, we focus on regimes that go beyond the widely adopted DeepSeek-R1- and Search-R1-style baseline: a simple, effective two-stage process with an optional cold start followed by reinforcement learning to align the model with long-horizon objectives. On top of this baseline, recent work introduces significant innovations to handle increasing task complexity and tool-augmented environments.

\paragraph{The Standard Agentic RL Pipeline}
Before diving into per-paper innovations, we briefly outline the common RL training pipeline used in deep research agents, using Search-R1 as the canonical reference. First, some works apply a cold start (e.g., supervised fine-tuning or rejection sampling fine-tuning (SFT/RSFT)) to teach interface compliance (e.g., tool schemas) and stabilize early rollouts; others skip this step. During RL training, the policy is given a complex query and generates a trajectory $\tau$ with interleaved reasoning and tool use. A prompt template enforces a parseable, ReAct-style~\citep{yao2023react} structure with explicit tags, i.e., \texttt{$<$think$>$...$<$/think$>$} for reasoning, \texttt{$<$search$>$...$<$/search$>$}to trigger retrieval, \texttt{$<$information$>$...$<$/information$>$} for injected results, and \texttt{$<$answer$>$...$<$/answer$>$} for the final response. This supports multi-turn search-and-reason loops until the model answers or an action budget is exhausted.

\begin{tcolorbox}[colback=rliableblue!10!white,
                  colframe=black,
                  boxrule=1pt,
                  boxsep=2pt,
                  top=3pt,
                  bottom=3pt,
                  left=2pt,
                  right=2pt]
\textbf{[Search-R1 Template]} \\
Answer the given question. You must conduct reasoning inside \textcolor{blue}{\texttt{$<$think$>$}} and \textcolor{blue}{\texttt{$<$/think$>$}} first every time you get new information. After reasoning, if you find you lack some knowledge, you can call a search engine by \textcolor{blue}{\texttt{$<$search$>$}} query \textcolor{blue}{\texttt{$<$/search$>$}}, and it will return the top searched results between \textcolor{blue}{\texttt{$<$information$>$}} and \textcolor{blue}{\texttt{$<$/information$>$}}. You can search as many times as you want. If you find no further external knowledge needed, you can directly provide the answer inside \textcolor{blue}{\texttt{$<$answer$>$}} answer \textcolor{blue}{\texttt{$<$/answer$>$}} without detailed illustrations. For example, \textcolor{blue}{\texttt{$<$think$>$}} xxx \textcolor{blue}{\texttt{$<$/think$>$}}. 
Question: \textcolor{red}{question}.
\end{tcolorbox}

Many pipelines combine an outcome reward (i.e., final answer correctness) with a small format reward to encourage well-formed traces. While most works adopt a composite reward of outcome and format, some work (including Search-R1) uses outcome-only, rule-based rewards (e.g., EM on extracted answers) and shows that this suffices under their setup. Optimization is typically PPO or GRPO with KL regularization to a reference policy. Crucially, in tool-augmented optimization, tokens generated by tools are masked out so that gradients (and KL) are computed only on model-generated tokens, thereby stabilizing training with interleaved tool use. Advantages come from GAE in PPO or group baselines in GRPO.

Putting this together, the baseline loop is: (i) optional cold start; (ii) templated rollouts with explicit tags and an action budget; (iii) tool returns injected into the context; (iv) reward computation (at least outcome; optional format); (v) policy update via PPO/GRPO with KL to a reference and masking of non-policy tokens. This yields a policy that learns the reasoning skeleton and when/how to invoke tools, with stability anchored by the reference-KL and token masking.

\subsubsection{Cold-Start Choices}
While half of the papers in Table~\ref{tab:rl-regime}, \ref{tab:rl-reward} and \ref{tab:rl-multimodal} skip a cold start for simplicity, the rest adopt a two-stage pipeline (cold start with SFT/RSFT followed by RL training), and report clear benefits: improved early-stage stability (e.g., avoiding reward collapse) and faster, more sample-efficient convergence by quickly mastering formats~\citep{li2025websailor,tao2025webshaper,dong2025ARPO,tan2025ragr1}. For example, \textit{RAG-R1} \citep{tan2025ragr1} applies SFT before RL and argues that SFT is critical for leveraging both internal and external knowledge; \textit{WebSailor} \citep{li2025websailor} shows that a modest RSFT cold start is indispensable for complex web tasks because early RL signals are extremely sparse and learning is slow due to multi-turn, heavy tool use; Their ablation comparing direct RL vs. RFT $\rightarrow$ RL finds that the cold-started model converges to much higher final performance. In the same spirit, \textit{ARPO} \citep{dong2025ARPO} adopts a cold start before RL training explicitly to mitigate reward collapse during the initial RL phases. Benefits also appear in other modalities, e.g., video (\textit{Ego-R1}), visual (\textit{WebWatcher}), and table reasoning (\textit{Reasoning-Table}). Despite these trends, the optimal extent and duration of any cold start prior to RL remain open questions~\citep{shi2025pangu}.

\subsubsection{Curriculum Over the Pipeline}
Another line of research strengthens the training regime by incorporating curriculum over the standard pipeline. \textit{EVO-RAG} \citep{ji2025curriculum} introduces a two-stage curriculum: \emph{discovery} to encourage broad, diverse querying, followed by \emph{refinement} that steers the agent toward concise, targeted queries for evidence-grounded answers. \textit{Writing-RL} \citep{lei2025writing} generalizes this idea to multi-stage curricula and adds \emph{margin-aware data selection}, which estimates learning headroom (the gap between the policy’s output and the strongest reference) to prioritize high-potential samples; it further reports consistent gains of curriculum over non-curriculum training. Building on these ideas, \textit{EvolveSearch} \citep{zhang2025evolvesearch} applied curriculum learning iteratively to both SFT and RL training, achieving further performance improvements. In all cases, the optimizer (e.g., PPO/GRPO) remains standard; the innovation lies in how data are staged, scheduled, and escalated in difficulty throughout training.

\subsubsection{Optimizers in Practice}
In terms of RL optimizers, most recent papers adopt GRPO as the primary algorithm (e.g., \citet{chen2025researchlearningreasonsearch,shi2025pangu,singh2025artist,java2025frugalrag,zhang2025evolvesearch,gao2025asearcher,fan2025ssrl}), reporting strong benchmark results. A second line of work uses PPO as the main trainer (e.g., \citet{jin2025search,zhou2025mem1,tan2025ragr1,jiang2025s3}). Studies that compare both (e.g., \citet{jin2025search,sun2025zerosearch,zhao2025rsearchempoweringllmreasoning}) generally find that GRPO has simpler mechanics and converges with fewer updates, whereas PPO tends to provide greater training stability under noisy, long-horizon returns; final reward quality is often comparable. Beyond the GRPO/PPO choice, several alternatives appear. \textit{R1-Searcher} \citep{song2025r1} employs REINFORCE++, observing that GRPO generalizes better out-of-domain while REINFORCE++ achieves higher data efficiency and stronger in-domain scores. \textit{ZeroSearch} \citep{sun2025zerosearch} uses vanilla REINFORCE, reporting gains over PPO/GRPO within its pipeline. \textit{WebSailor} \citep{li2025websailor} introduces Duplicating Sampling Policy Optimization (DUPO), an improvement over DAPO~\citep{yu2025dapo}, that adds two dynamic sampling strategies: (i) a pre-training filter to drop trivially easy queries (all rollouts correct), and (ii) in-training duplication of queries with non-zero return variance. DUPO keeps batches informative without extra sequential rollouts and reports a $\sim 2$–$3\times$ speed-up over DAPO’s dynamic sampling. Orthogonal to the optimizer itself, \textit{ARPO} \citep{dong2025ARPO} augments trajectory-level RL with entropy-triggered partial rollouts at tool steps—branching only when post-tool token entropy spikes—together with segment-aware advantage attribution so shared prefixes and branched continuations receive different credits; this targets exploration where tool feedback raises uncertainty and improves tool efficiency without inflating rollout cost. 

\subsubsection{Context Control}
Existing research also targets limitations of current agentic training pipelines, such as high memory consumption and limited adaptability \citep{zhou2025mem1,shi2025searchrefinethinkautonomous,li2025websailor}. \textit{MEM1} \citep{xu2025amemagenticmemoryllm} observes that prompt length grows across multi-turn search and addresses this by replacing accumulated history with a compact internal state that is rewritten at every step; earlier turns are pruned, so the working context remains near-constant while still carrying forward the necessary reasoning and tool observations. \textit{AutoRefine} \citep{shi2025searchrefinethinkautonomous} tackles the issue with a search-and-refine loop that interleaves retrieval with explicit evidence distillation: after each search, long documents are compressed into short ``refine notes'' that preserve only the crucial evidence to be used in subsequent steps, preventing prompt growth without sacrificing fidelity. \textit{WebSailor} \citep{li2025websailor} tackles the context issue from another angle: it employs a lightweight rejection-sampling fine-tuning step to reconstruct verbose reasoning and tool-use traces into concise, consistent, action-oriented sequences, thereby improving trace quality and reducing prompt bloat without sacrificing fidelity. Research on agent architecture and coordination designs also recognizes this issue (see Section~\ref{ch:architecture}). In brief, these systems decouple a domain-agnostic planner from domain-specific executors, and aggregate/distill intermediate results before returning them to the planner, keeping the planner’s working context short, clean, and stable over long horizons.

\subsubsection{Learning Search Necessity}
Recent work makes ``when to search'' a learned decision so agents lean on parametric knowledge and retrieve only when needed. \textit{R1-Searcher++} \citep{song2025r1searcher++} separates internal vs. external traces in SFT (with masked external text), then uses outcome-based RL plus a group penalty and a memorization module that rewrites retrieved facts into internal traces—reducing future lookups. \textit{FrugalRAG} \citep{java2025frugalrag} runs a lightweight SFT to learn high-recall exploration from ReAct rollouts, then applies GRPO to learn an explicit STOP action that trades additional queries against confidence, adapting test-time compute per question. \textit{IKEA} \citep{huang2025IKEA} builds a knowledge-boundary-aware policy: prompts bias toward internal recall, GRPO uses rewards that favor correct answers with fewer retrievals, and a balanced mix of internal-answerable vs. external-required examples teaches the agent to search only when its own knowledge is insufficient. Complementarily, SSRL trains a self-search behavior (internal retrieval within the trajectory) via RL and then swaps to real search at inference, effectively learning to defer external queries until necessary~\citep{fan2025ssrl}.

\subsubsection{Cost \& Latency-Aware Training}
To rein in the cost of live retrieval and keep inference responsive, two complementary strategies have emerged. Simulation-based training removes web/API calls during RL: \textit{ZeroSearch} \citep{sun2025zerosearch} replaces the real engine with an LLM-based search simulator, enabling cheap, unlimited rollouts while simulating real-web dynamics by employing a noise curriculum, cutting API spend by orders of magnitude without hurting QA quality. \textit{SSRL} \citep{fan2025ssrl} pushes this idea inside the model, structuring a self-search loop within the trajectory and optimizing it with GRPO; training remains fully in-simulation yet transfers well to real search, yielding substantial speedups. Orthogonally, \textit{RAG-R1} \citep{tan2025ragr1} reduces per-question latency by issuing multi-query searches in parallel and fuses evidence, teaching the agent when and how to search efficiently, thus reducing retrieval rounds and end-to-end time without sacrificing EM.

\subsubsection{Asynchronous Rollout and Training}
\textit{ASearcher} \citep{gao2025asearcher} implements a fully asynchronous actor–learner design that decouples rollout generation from policy updates and tolerates stragglers, enabling very long-horizon trajectories with heavy tool use (search+browse) without idling the learner. Summarization and evidence aggregation are learned end-to-end within the RL loop (GRPO), while a simple dynamic filter removes zero-signal prompts. Although the contribution is primarily systems-level, these choices alter the feasible training regime—unlocking stable, sample-efficient long-context training—so we place it here; for engineering details and broader agentic RL training frameworks, see Section~\ref{ch4:infra}.

\subsubsection{Discussion}
Across recent work, the baseline pipeline has solidified: optional cold start to teach interfaces, templated rollouts with explicit tool tags and action budgets, outcome(-plus-format) rewards, and PPO/GRPO with KL-to-reference while masking tool-return tokens. On top of this, three themes stand out. First, \emph{stability and efficiency}: cold starts and curricula speed convergence and prevent early reward collapse; token masking and reference-KL anchor learning despite interleaved tool text. Second, \emph{data/compute shaping}: curricula, dynamic sampling (e.g., duplication/filters), and entropy-triggered branching focus updates where uncertainty and learning headroom are highest. Third, \emph{cost/latency control and search necessity}: context-control mechanisms (state rewriting, evidence distillation) and policies that learn when to search (including STOP actions and knowledge-boundary awareness) reduce retrieval rounds, while simulators and parallel query strategies cut training/inference costs without sacrificing answer quality. Optimizer choice (GRPO vs. PPO vs. REINFORCE-family) largely trades simplicity and speed for robustness under long-horizon, noisy returns; final quality is often comparable when the surrounding regime (masking, KL, curricula) is well-tuned.

We list three open questions in this area: (i) Cold start and curriculum scheduling: How should we automatically decide \emph{when} to stop SFT/RSFT, advance curriculum phases, or switch difficulty to maximize sample efficiency without overfitting? (ii) Optimizer–tool interaction: What principled criteria pick PPO/GRPO/REINFORCE++ under partial, delayed, and segment-wise credit with tool boundaries—and can segment-aware advantage attribution be unified with KL control for stronger stability? (iii) Cost-aware objectives: How do we train truly multi-objective policies that optimize accuracy and explicit budgets (latency, queries, tokens), with guarantees on test-time compute allocation (reasoning depth, retrieval rounds)?

\subsection{Reward Design and Credit Assignment} \label{ch3.2:reward}
Reward design and credit assignment are central to RL training. While training regime and optimization structure dictate how learning unfolds, reward strategies determine which behaviors are reinforced and how credit is allocated across complex trajectories.

\begin{table*}[ht]
  \footnotesize
  \centering
  \caption{Summary of papers in Section \ref{ch3.2:reward}. 
  }
  \begin{tabularx}{\textwidth}{
  @{}p{2.8cm}@{\hspace{0.2cm}}p{3.6cm}
  @{\hspace{0.3cm}}>{\centering\arraybackslash}p{1.0cm}
  p{3cm}@{\hspace{0.2cm}}p{2.5cm}@{} 
  }
    \toprule
    \textbf{Work} & \textbf{Policy Model} & \textbf{Cold Start?} & \textbf{Reward Type} & \textbf{RL Optimizer}\\
    \midrule
    \textit{s3}\newline\citep{jiang2025s3}   &Qwen2.5-7B-Instruct   & \xmark & Gain\_Beyond\_RAG &PPO \\
    \addlinespace
    \textit{AutoRefine}\newline\citep{shi2025searchrefinethinkautonomous}   &Qwen2.5-3B-Base/Instruct   & \xmark & Outcome\newline+Retrieval\_Quality &GRPO \\
    \addlinespace
    \textit{MT-GRPO}\newline\citep{zeng2025MTGRPO}   &Qwen2.5-7B-Base   & \xmark & Outcome+Format\newline+Retrieval\_Quality &MT-GRPO\newline (GRPO-variant)\\
    \addlinespace
    \textit{IKEA}\newline\citep{huang2025IKEA}   &Qwen2.5-7B-Base/Instruct      & \xmark & Outcome+Format\newline+Retrieval\_Necessity &GRPO \\
    \addlinespace
    \textit{ARTIST}\newline\citep{singh2025artist}   &Qwen2.5-14B-Instruct   & \xmark & Outcome+Format\newline +Tool\_Execution &GRPO \\
    \addlinespace
    \textit{R1-Searcher++}\newline\citep{song2025r1searcher++}   &Qwen2.5-7B-Instruct      & \cmark & Outcome+Format\newline+Retrieval\_Necessity &Reinforce++ \\
    \addlinespace
    \textit{StepSearch}\newline\citep{wang2025stepsearch}   &Qwen2.5-7B-Base/Instruct      & \xmark & Outcome+Format\newline+(InfoGain-Redundancy) &StePPO\newline (PPO-variant) \\
    \addlinespace
    \textit{$O^{2}$-Searcher}\newline\citep{mei2025o2searcher}   &Qwen2.5-3B-Instruct      & \cmark & Outcome+Format\newline+LongText\_Metrics &GRPO \\
    \addlinespace
    \textit{R-Search}\newline\citep{zhao2025rsearchempoweringllmreasoning}   &Qwen2.5-7B-Instruct      & \xmark & Outcome+Format\newline+Evidence &PPO,\newline GRPO \\
    \bottomrule
  \end{tabularx}
  \label{tab:rl-reward}
\end{table*}

\subsubsection{Outcome-level Rewards}
Most studies frame multi-turn interaction as a bandit problem \citep{zeng2025reinforcing}, evaluating performance at the terminal step (e.g., answer correctness and formatting). We organize outcome-level rewards into: (i) classical metrics, (ii) composition strategies, and (iii) novel outcome-level signals that target search/evidence/efficiency beyond correctness and format.

\paragraph{Classical metrics}
Terminal rewards traditionally score the final response with (i) exact match (EM) / F1 for correctness, (ii) LLM-as-a-judge graders for holistic quality, and (iii) task metrics such as NDCG in retrieval-style tasks. Representative examples include rule-based paper-query alignment in PaSa \citep{he-etal-2025-pasa}, LLM-graded writing quality in Writing-RL \citep{lei2025writing}, loose/strict LLM graders in DeepDiver \citep{shi2025pangu}, and task-specific metrics (e.g., NDCG) in SAGE \citep{wang2025sage} and VRAG-RL \citep{wang2025vragrl}.

\paragraph{Composition strategies}
Outcome signals are typically composed via: (i) additive (weighted sums) that combine format and correctness (e.g., \citealp{mei20252, song2025r1searcher++, zhao2025rsearchempoweringllmreasoning}); (ii) hierarchical/conditional schemes that gate the final reward on pass/fail conditions to avoid over-penalizing partially correct reasoning \citep{ren2025knowrlexploringknowledgeablereinforcement, shi2025search, huang2025IKEA, lei2025reasoningtable}; and (iii) group-relative bonuses that compare rollouts within a sampled group (e.g., \citealp{song2025r1searcher++}). These choices control stability and what behaviors are emphasized at the terminal step.

\paragraph{Novel outcome-level signals}
Beyond standard answer/format scoring, several works introduce outcome-level rewards that target search, evidence, and tool economy directly:
\begin{itemize}
    \item \textbf{Gain-Beyond-RAG (GBR; \cite{jiang2025s3}).} Rewards the \emph{delta} in generation accuracy when using the agent’s retrieved context versus a naive top-$k$ RAG baseline, attributing improvement specifically to the searcher.
    \item \textbf{Cross-model evidence utility.} \textit{R-Search} \citep{zhao2025r} treats agent-written evidence as a self-contained bundle: a \emph{frozen external LLM} must answer correctly from this evidence; the reward reflects the downstream recoverability of the gold answer.
    \item \textbf{Knowledge-boundary shaping.} \textit{IKEA} \citep{huang2025IKEA} designs a hierarchical terminal signal that discourages redundant external calls when the model already knows the answer, while lightly encouraging retrieval when it does not, aligning usage with actual need.
    \item \textbf{Group-relative efficiency.} \textit{R1-Searcher++} \citep{song2025r1searcher++} grants a bonus to trajectories that use the \emph{fewest} external calls among correct rollouts in a sampled group, thereby incentivizing internalization and thriftiness.
    \item \textbf{Query diversity.} \textit{$O^{2}$-Searcher} \citep{mei2025o2searcher} adds a diversity-aware outcome term that promotes non-duplicative queries under a controlled budget, mitigating mode collapse in query generation.
    \item \textbf{Refinement/coverage.} \textit{AutoRefine} \citep{shi2025searchrefinethinkautonomous} gives credit when the final refined knowledge state explicitly contains the required gold-answer components, rewarding evidence consolidation beyond mere format/correctness.
\end{itemize}

In practice, these novel signals are used alongside classical outcome terms, e.g., \textit{O$^2$-Searcher}, \textit{R1-Searcher++}, and \textit{R-Search} retain format/answer components while augmenting them with diversity, group-relative, or evidence-utility rewards.

\subsubsection{Step-level Rewards}
Outcome-level rewards treat the entire decision trajectory as a single decision step, thereby overlooking the multi-turn structure of the task. This limitation hinders the model’s ability to learn robust reasoning chains. To address this, researchers \citep{zeng2025reinforcing, wang2025stepsearch, liu2025visual} have explored incorporating step-level rewards alongside outcome-level rewards, enabling more fine-grained credit assignment. 

\paragraph{Tool/execution \& presence checks}
To expose fine-grained supervision within a trajectory, several works attach per-turn rewards to how a tool is used and what it returns. \textit{MT-GRPO}~\citep{zeng2025reinforcing} grants a small reward when a tool call is well-formed and successfully executed, and another when the retrieved snippets contain any gold answer span, thereby reinforcing correct invocation and retrieval sufficiency at the moment they occur. \textit{ARTIST}~\citep{singh2025artist} further instantiates this pattern with a Function Reward for issuing the correct sequence of function calls and a State Reward for correct state tracking during multi-turn tool use, i.e., explicit step-level execution checks.

\paragraph{Information gain vs.\ redundancy}
Instead of merely checking presence, \textit{StepSearch}~\citep{wang2025stepsearch} shapes each turn by rewarding novel, useful evidence and penalizing repeats. Concretely, a step reward adds information gain, i.e., the marginal similarity improvement between the round’s retrieved documents and a reference set of gold evidence, minus a redundancy penalty that increases with overlap against earlier rounds, steering the agent toward diversified, high-yield hops.

\paragraph{Query-intent alignment}
Complementing evidence-level shaping, \textit{StepSearch}~\citep{wang2025stepsearch} adds a query-intent signal: overlap between the model’s generated query and reference sub-task keywords for that turn. This per-turn reward keeps search on-task (e.g., targeting the correct sub-question in a decomposition) and reduces drift, without waiting for the final answer to provide feedback.

\paragraph{Multimodal turn checks}
For multimodal agents, \textit{Visual-ARFT}~\citep{liu2025visual} extends turn-level shaping beyond text by scoring format quality, search/tool usage, and code quality at each step, providing dense feedback that trains the full perception–tool–reason loop rather than only the terminal output.

In practice, these step-level signals are combined with outcome-level rewards to provide dense supervision for multi-turn tasks; reward placement typically attaches step rewards at round boundaries and the terminal reward at the end of the trajectory.

\subsubsection{Credit Assignment}
Effective credit assignment is pivotal in multi-turn RL because the agent’s useful behaviors (e.g., deciding \emph{when} to search, \emph{what} to retrieve, and \emph{how} to answer) occur at different times; without careful credit flow, terminal rewards either over-credit the last step or under-credit key intermediate decisions.

\paragraph{Trajectory-level estimators}
Many work use terminal-only signals with standard policy-gradient optimizers—REINFORCE, GRPO, or PPO—plus a KL penalty to a frozen reference model to stabilize updates. In tool-augmented settings, implementations typically mask injected tokens (e.g., pasted documents) and compute loss only on action tokens (tool invocation, reasoning/answer tokens), so gradients do not spuriously pass through the evidence text \citep{song2025r1searcher++, zhao2025rsearchempoweringllmreasoning, huang2025IKEA}. This setup assigns a single trajectory-level advantage to the full sequence and is simple and robust, but can be slow to propagate credit to early, causally important steps.

\paragraph{Turn-level estimators}
To better align credit with the process structure, \textit{MT-GRPO}~\citep{zeng2025MTGRPO} computes per-turn advantages that blend immediate turn feedback with final-outcome advantages, improving learning for multi-turn tool use agents. Concretely, early turns receive credit from both their own turn rewards (e.g., search quality) and a portion of the outcome signal, while later turns are outcome-focused, yielding finer attribution than trajectory-level estimators without changing the underlying RL optimizer.

\paragraph{Token/round placement and masking}
Even with standard optimizers like PPO/GRPO, \emph{where} rewards are attached matters. Process-aware implementations such as \textit{StepSearch} \cite{wang2025stepsearch} attach the terminal reward to the final tokens and attach step rewards at the end of each search round, allowing GAE to propagate credit locally through each round while still reflecting downstream success. Step-level execution rewards in \textit{ARTIST} \citep{singh2025artist} are similarly attached at the tool-call boundary (function and state checks), providing immediate feedback on action quality within the turn. As above, action-token-only loss and observation masking are maintained to avoid leaking gradients through retrieved content. 

\subsubsection{Discussion}
Overall, verifiable terminal rewards remain the anchor, but how they are composed (additive vs. conditional vs. group-relative) and where credit flows (trajectory vs. turn vs. token/round) affect stability and sample efficiency. Novel outcome signals target search, evidence quality, and tool economy; step-level shaping helps, but dense process rewards can trigger reward hacking or tool avoidance, making outcome-first designs with tight action budgets more stable.

Key open questions include: (i) how to compose and schedule multi-objective rewards (e.g., correctness, GBR, diversity, evidence utility) in a principled, anti-hacking way; (ii) how to build robust, low-cost verifiers that replace expensive or easily gamed LLM judges without sacrificing alignment; (iii) how to deliver causal, low-variance credit to search/tool choices via turn/token-level methods; and (iv) how to learn budget- and risk-aware policies that trade off accuracy, latency, and tool cost without inducing tool avoidance.

\subsection{Multimodal Research Agents} \label{ch3.3:multimodal}
As deep research agents expand beyond text, a central question is how to build agents whose policy natively perceives and reasons over multiple modalities (e.g., vision and language). This section surveys end-to-end multimodal models (typically VLMs) that perform iterative perception–reasoning cycles. We explicitly exclude works that outsource multimodality to off-the-shelf OCR/vision modules, code interpreters, or retrieval heuristics. Instead, we focus on models that ingest visual evidence directly and produce grounded reasoning traces and answers within a unified multimodal token space. Despite recent progress, multimodal deep research agents remain comparatively early-stage relative to text-only LLM agents.

\begin{table*}[ht]
  \footnotesize
  \centering
  \caption{Summary of papers in Section \ref{ch3.3:multimodal}.}
  \begin{tabularx}{\textwidth}{
  @{}p{2.8cm}@{\hspace{0.2cm}}p{3.8cm}
  @{\hspace{0.3cm}}>{\centering\arraybackslash}p{1.0cm}
  p{3cm}@{\hspace{0.2cm}}p{2.5cm}@{} 
  }
    \toprule
    \textbf{Work} & \textbf{Policy Model} & \textbf{Cold Start?} & \textbf{Reward Type} & \textbf{RL Optimizer}\\
    \midrule
    \textit{Visual-ARFT}\newline\citep{liu2025visual}   &Qwen2.5-VL-7B-Instruct   & \xmark & Outcome+Format\newline+Retrieval\_Quality &GRPO \\
    \addlinespace
    \textit{VRAG-RL}\newline\citep{wang2025vragrl}   &Qwen2.5-VL-7B-Instruct  & \cmark & Outcome+Format\newline+Retrieval\_Quality &GRPO \\
    \addlinespace
    \textit{WebWatcher}\newline\citep{geng2025webwatcher}   &Qwen2.5-VL-32B-Instruct          & \cmark & Outcome+Format &GRPO \\
    \addlinespace
    \textit{MMSearch-R1}\newline\citep{wu2025mmsearch}   &Qwen2.5-VL-7B-Instruct          & \xmark & Outcome+Format &GRPO \\
    \addlinespace
    \textit{V-ToolRL}\newline\citep{su2025vtoolrl}   &Qwen2-VL-2B-Instruct          & \cmark & Outcome &GRPO \\
    \addlinespace
    \textit{VTool-R1}\newline\citep{wu2025vtool}   &Qwen2.5-VL-32B-Instruct          & \xmark & Outcome &GRPO \\
    
    \bottomrule
  \end{tabularx}
  \label{tab:rl-multimodal}
\end{table*}

\subsubsection{Multimodal Action–Observation Interface}
Across recent multimodal deep research agents, the RL optimizer is essentially unchanged (GRPO/PPO with a KL reference, masking tool-generated tokens, and optimizing only final-answer tokens); the real novelty lies in the \emph{state/action/observation} contract \citep{liu2025visual,wang2025vragrl,geng2025webwatcher,wu2025mmsearch,su2025vtoolrl,wu2025vtool}, as shown in \Cref{tab:rl-multimodal}. In multimodal settings, \emph{perception becomes an action}: the policy issues visual operations-(i) region crop/zoom reprojected to raw pixels for true DPI on dense charts/tables, (ii) edit-then-reason steps (highlight/mask/box) with dual-image conditioning $(I \oplus I')$, and (iii) lightweight image code (rotate/denoise/brighten)-and then \emph{re-conditions} on their consequences, turning one-shot perception into a controllable evidence-update loop \citep{wang2025vragrl,wu2025vtool,liu2025visual}. This works only with observation engineering for long contexts and noise control (e.g., raw-pixel crops over thumbnails; reader$\rightarrow$summary transforms for webpages) and with unified, tagged action schemas (visual acts and web/text search under \texttt{<action>…</action>}) coupled to tight action budgets for stability \citep{wu2025mmsearch,su2025vtoolrl,geng2025webwatcher}.

\paragraph{Observation Engineering \& Grounding} 
Because images are high-entropy, agents must see \emph{usable} evidence. Effective choices include (a) re-encoding raw-pixel crops (not encoder-space thumbnails) to recover small text and chart detail \citep{wang2025vragrl}, (b) dual-image reinsertion after edits to externalize attention \citep{wu2025vtool}, (c) strict, typed tool-return schemas with controller-level normalization/caching \citep{su2025vtoolrl}, and (d) reader$\rightarrow$summary webpage processing to suppress boilerplate \citep{wu2025mmsearch}. Trajectory quality gates—schema checks, step-consistency filters, and \emph{entity obfuscation paired with retrieved images} to force genuine visual grounding—prevent ``answer-from-prior'' shortcuts \citep{geng2025webwatcher}.

\paragraph{Learning Evidence Necessity}
In multimodal agents, the classic ``when to search'' problem generalizes to \emph{evidence necessity}: choosing which modality to query first (image vs.\ text), whether to perform perception actions (crop/zoom, highlight/mask, lightweight image code) before or instead of retrieval, when to hand off from image search to text reading, and when to stop. Recent agents implement this via a unified action space (including \textsc{no-op}/\textsc{finish}) and outcome-centric rewards augmented with light, modality-aware efficiency signals—e.g., a \emph{search penalty} for thrift in web calls \citep{wu2025mmsearch}, trajectory-level \emph{image-retrieval ranking} (NDCG) to surface the right visual earlier \citep{wang2025vragrl}, and a binary \emph{executable-image-code} check to encourage safe preprocessing exploration \citep{liu2025visual}. Datasets are balanced across \emph{search-/vision-required} vs.\ -free items and filtered with strict schema/consistency gates to teach selectivity rather than reflexive tool use; the optimizer remains standard while evaluation couples quality with thrift (search ratio, perception-action counts, image NDCG, executable-code rate) \citep{wu2025mmsearch,wang2025vragrl,liu2025visual}.

\subsubsection{Discussion}
Recent progress in multimodal agents is driven less by new optimizers and more by perception-as-action, engineered observations, and a lean RL recipe (GRPO/PPO+KL with tool-token masking and outcome-first rewards). Together, these techniques transform brittle one-shot reads into controllable evidence updates. They improve stability through design choices such as raw-pixel crops, dual-image reconditioning, and reader-to-summary transforms, while also instilling evidence necessity—deciding which modality to use and whether to perceive before retrieving—in order to curb over-search and strengthen grounding.

We identify three promising directions for multimodal deep research agents: 
(i) tracing performance gains back to specific perception steps or image regions; 
(ii) multi-image/multi-page reasoning at scale (PDFs, dashboards) without exploding context; 
(iii) developing standardized \emph{process+thrift} benchmarks and reporting (action budgets, masking policy, cache/rate limits) for apples-to-apples comparison. 
\section{Agentic RL Training Frameworks} \label{ch4:infra}
Deep research agents learn through long, tool-using interactions; making them trainable with RL is therefore a systems problem. This section surveys open-source infrastructure for agentic RL released in 2025 and is organized around three questions: (1) what bottlenecks currently limit training, (2) what design patterns and system mechanisms recent frameworks contribute to address them, and (3) how to choose and compose frameworks in practice.

\begin{table*}[ht]
  \footnotesize
  \caption{Open-source agentic RL training frameworks for LLMs (listed alphabetically).}
  \centering
  \begin{tabularx}{\textwidth}{
    @{}p{2.6cm} p{9.3cm} p{2.5cm}@{}}
    \toprule
    \textbf{Framework} & \textbf{Key features} & \textbf{GitHub} \\
    \midrule
    \textit{Agent Lightning}\newline \citep{luo2025agentlightning} 
      & Trainer–agent disaggregation (server–client);\newline unified transition/MDP interface;\newline LightningRL credit assignment for multi-turn traces;\newline telemetry \& automatic intermediate rewards (AIR) 
      & \href{https://github.com/microsoft/agent-lightning}{\faGithub\ repo} \\
    
    \addlinespace
    \textit{AReaL}\newline \citep{fu2025areal}        
      & Fully asynchronous actor-learner with high MFU;\newline interruptible/cancelable decoding;\newline staleness-aware PPO for mixed-policy batches;\newline dynamic packing \& parallel reward service 
      & \href{https://github.com/inclusionAI/AReaL}{\faGithub\ repo} \\

    \addlinespace
    \textit{OpenR}\newline \citep{wang2024openr}        
      & PRM-centric process supervision;\newline PRM-guided decoding/test-time compute;\newline unified data + online/offline RL stack;\newline gym-style MDP with PPO 
      & \href{https://github.com/openreasoner/openr}{\faGithub\ repo} \\

    \addlinespace
    \textit{rLLM}\newline \citep{rllm2025}         
      & Async, OpenAI-compatible rollout engine;\newline stepwise GRPO variants for long horizons;\newline observation masking (action-only loss);\newline verl-backed distributed PPO/GRPO 
      & \href{https://github.com/rllm-org/rllm}{\faGithub\ repo} \\

    \addlinespace
    \textit{ROLL}\newline \citep{wang2025roll}         
      & Single controller + workerized Actor/Critic/Env/Reward;\newline sample-level scheduler \& async reward workers;\newline AutoDeviceMapping for heterogeneous clusters;\newline 5D parallelism with vLLM/SGLang backends 
      & \href{https://github.com/alibaba/ROLL}{\faGithub\ repo} \\

    \addlinespace
    \textit{SLIME}\newline \citep{2025SLIME}        
      & SGLang-native serving (router, PD/EP);\newline Megatron-native trainer, large-MoE ready;\newline async/decoupled rollout–train via Ray;\newline abort in-flight \& frequent weight sync 
      & \href{https://github.com/THUDM/slime}{\faGithub\ repo} \\

    \addlinespace
    \textit{Verifiers}\newline \citep{2025verifiers}    
      & Protocol-first MultiTurn/Tool environments;\newline OpenAI-compatible with vLLM rollout;\newline built-in GRPO trainer (Accelerate/DeepSpeed/LoRA);\newline rubric/judge-based, tool-aware rewards 
      & \href{https://github.com/willccbb/verifiers}{\faGithub\ repo} \\

    \addlinespace
    \textit{verl}\newline \citep{sheng2025verl}    
      & HybridFlow single-/multi-controller APIs;\newline 3D-HybridEngine for zero-redundancy resharding across train $\leftrightarrow$ gen;\newline scalable 3D/ZeRO/FSDP/Megatron parallelism with auto device mapping;\newline high-throughput RLHF/RLAIF pipelines. 
      & \href{https://github.com/volcengine/verl}{\faGithub\ repo} \\
    \bottomrule
  \end{tabularx}
  \label{tab:rl-frameworks}
\end{table*}

\subsection{Bottlenecks \& challenges in agentic RL training}
We identify five recurring bottlenecks and challenges in current agentic RL training:
\begin{itemize}
  \item \textbf{Rollout throughput and latency.} Long, multi-turn, tool-using episodes stall GPUs (e.g., longest-sample tails, synchronization barriers), so experience collection often dominates end-to-end cost \citep{fu2025areal}.
  
  \item \textbf{Policy staleness and unstable credit assignment.} Asynchronous or mixed-policy batches violate standard PPO-style assumptions, while sparse/delayed rewards over long trajectories increase variance and hinder stable improvement; staleness-aware objectives and stepwise credit assignment are typically required \citep{fu2025areal}.
  
  \item \textbf{Large-scale orchestration.} Switching the same model between training and generation, mapping TP/PP/DP (and MoE/EP) parallelism, and moving tensors without redundancy are non-trivial at cluster scale; efficient, zero-redundancy train$\leftrightarrow$gen transitions remain a core systems challenge \citep{sheng2025verl}.
  
  \item \textbf{Heterogeneous agent runtimes.} Production agents (e.g., LangChain/AutoGen/custom stacks) are tightly coupled to tools and execution logic, making it difficult to ``drop in'' RL without refactoring; clean trainer–agent disaggregation and unified transition interfaces are needed \citep{luo2025agentlightning}.
  
  \item \textbf{Outcome-only supervision is weak for multi-step reasoning.} Pure outcome reward signals under-specify long-horizon behavior; process-aware supervision (e.g., PRMs), verifiable/tool-aware rewards, and structured protocols are needed to shape trajectories \citep{wang2024openr}.
\end{itemize}

\subsection{What the frameworks contribute (methods \& features)}
To address the aforementioned challenges, the mainstream open-source training frameworks propose the following methods and features. 

\paragraph{Addressing rollout throughput and latency.}
To make rollouts fast and elastic, recent systems raise sampling throughput by adopting fully asynchronous actor–learner designs that remove batch-wide waits. \textit{AReaL} \citep{fu2025areal} proposes interruptible/cancelable decoding together with staleness-aware PPO to stabilize learning while boosting Model Flops Utilization (MFU). Serving-native RL sampling further helps: \textit{SLIME} \citep{2025SLIME} binds SGLang \citep{zheng2024sglang} (prefill–decode disaggregation and expert parallelism) to Megatron training, exposes an OpenAI-compatible router, and adds sampler features such as abort-in-flight tailored to dynamic or oversampling recipes. Complementarily, \textit{ROLL} \citep{wang2025roll} contributes sample-level schedulers and dedicated environment/reward workers to coordinate per-sample lifecycles, asynchronous rewards, and tool sandboxes across large clusters.

\paragraph{Bridging heterogeneous agent runtimes.}
\textit{Agent Lightning} \citep{luo2025agentlightning} reduces refactoring costs for production agents by introducing trainer–agent disaggregation with a unified transition interface, allowing existing stacks to be wrapped and treated as MDPs without code rewrites. In parallel, \textit{Verifiers} \citep{2025verifiers} adopts a protocol-first design with OpenAI-compatible endpoints, enabling the same module to support evaluation, data generation, and RL, and providing a lightweight GRPO path for multi-turn tool use.

\paragraph{Stabilizing credit assignment \& process supervision.}
For long-horizon behavior, \textit{rLLM} introduces hierarchical/stepwise GRPO variants that propagate or group advantages across steps, and \textit{Agent Lightning}'s LightningRL adds explicit credit assignment over multi-turn traces \citep{rllm2025, luo2025agentlightning}. \textit{OpenR} \citep{wang2024openr} integrates Process Reward Models (PRMs) and guided decoding to inject step-level signals into both training and test-time search, while \textit{Verifiers} \citep{2025verifiers} supplies rubric/judge-based, tool-aware rewards with parsers that make multi-criteria, process-aware supervision practical.

\paragraph{Systems co-design for scale}
\textit{verl} \citep{sheng2025verl} provides a hybrid single-/multi-controller programming model and a 3D-HybridEngine for zero-redundancy resharding across train$\leftrightarrow$gen, yielding substantial throughput gains across RLHF-style recipes. \textit{ROLL} \citep{wang2025roll} standardizes a single controller with parallel Actor/Critic/Reward/Env workers and AutoDeviceMapping, integrating Megatron-Core/FSDP for training and vLLM/SGLang for serving. \textit{SLIME} \citep{2025SLIME} packages large-MoE Megatron examples and server-managed SGLang pools behind one endpoint, easing frontier-scale agentic runs and simplifying deployment pathways.

\paragraph{Ecosystem-level conveniences}
Beyond core bottlenecks, frameworks add observability and shape signals to improve diagnosability and training efficacy. For example, \textit{Agent Lightning} \citep{luo2025agentlightning} integrates telemetry and automatic intermediate rewards into the runtime. Meanwhile, unified, end-to-end stacks are becoming common: \textit{rLLM} \citep{rllm2025} combines agents, environments, and stepwise GRPO; \textit{OpenR} \citep{wang2024openr} and releases code, models, and data for PRM-centric workflows.

\subsection{How to choose (pragmatic guidance)}
When the goal is a training back end that ``just runs'' PPO/GRPO with minimal glue, verl is a pragmatic starting point: it is production-ready, flexible, and its HybridFlow APIs simplify scalable train/gen switching and parallelism. It fits teams that already have a sampler and primarily need a robust engine for training and generation orchestration. If raw rollout throughput on long, tool-using episodes is the primary constraint, AReaL and SLIME are purpose-built for high MFU and dynamic sampling: AReaL provides a research-grade asynchronous blueprint with staleness-aware updates, while SLIME's SGLang-native sampling (with aborts and frequent weight refresh) targets serving-side speed under Megatron training \citep{fu2025areal,2025SLIME}. ROLL is a good fit when per-sample scheduling, environment/reward workers, and heterogeneous cluster management need to live in one library \citep{wang2025roll}.

If you already operate a production agent and you prefer not to rewrite, Agent Lightning's server–client split and unified transition interface lets you keep existing agent logic and tools while retrofitting RL \citep{luo2025agentlightning}. Domains that demand process-aware rewards or verifier-style signals benefit from OpenR's PRMs and guided decoding for reasoning-centric research and from Verifiers' protocol/rubric tooling (judge- and tool-aware rewards) coupled with a lightweight GRPO trainer \citep{wang2024openr,2025verifiers}. For frontier-scale parallelism or MoE, verl offers zero-redundancy transitions across train $\leftrightarrow$ gen with flexible device mapping, while ROLL or SLIME become attractive when you want, respectively, baked-in orchestration or serving-native scaling \citep{sheng2025verl,wang2025roll,2025SLIME}. 

A blended path many teams use is to adopt verl (or ROLL) as the training spine, choose SGLang or vLLM for sampling, and bring Verifiers (or a custom PRM/judge) for rewards; if a mature agent stack already exists, Agent Lightning can attach RL without refactors, and if throughput remains the blocker, asynchronous sampling such as AReaL or SLIME can be swapped in to raise utilization.

\subsection{Discussion}
This section frames agentic RL training as a systems problem and surveys recent frameworks through three lenses: (i) core bottlenecks; (ii) architectural responses—asynchrony, distributed orchestration, serving-native sampling, trainer–agent disaggregation, process-aware supervision hooks, and training–generation co-design; and (iii) pragmatic composition choices for training.

Looking ahead, we highlight three open questions for framework development: (i) Safety and robustness in online rollouts: What sandboxing, fault isolation, and guardrails are needed for browser/code/tool interactions under continual learning? (ii) Orchestration, elasticity, and fault tolerance at scale. How should schedulers handle preemption, multi-tenant QoS, partial-trajectory reclaim/cancel, and weight-sync semantics? (iii) Reproducibility, observability, and reporting. What trace/telemetry standards enable deterministic replays under non-deterministic tools/web, and which common ``process+thrift'' metrics (e.g., MFU, cost-per-success, staleness histograms, tool budgets) should frameworks report by default for standardized evaluation?

\section{Agent Architecture \& Coordination} \label{ch:architecture}
While our survey centers on RL foundations for training, the way agents are architected and coordinated is equally critical in practice. End-to-end RL over a full hierarchical stack remains impractical today because rollouts are long and high-latency, credit assignment spans many interacting components, and current infrastructure struggles with deterministic retrieval, tool sandboxes, and large-scale judging. As a result, most deployments compose pre-trained (often RL-enhanced) models into hierarchical or multi-agent systems rather than training the entire workflow jointly.

This chapter surveys those deployment-oriented designs: hierarchical planners (Planner–Coordinator–Executors), multi-agent teams, expert routing/gating, and modular sub-agents with shared state. We focus on how coordination mechanisms (e.g., task decomposition, scheduling, message passing, etc) turn individually trained models into scalable, robust deep research systems. Our aim is to outline the patterns and trade-offs that make these architectures effective today, and to clarify how stronger RL-trained planners can later slot into them to extend capability without incurring full end-to-end training cost.

\subsection{Open-Source Architectures \& Coordination}
We analyze four well-maintained open-source deployment frameworks for deep-research systems listed in Table~\ref{tab:deep-research-frameworks}, focusing on how they coordinate planning, delegation, tool use, and report assembly. These projects are not agentic RL training stacks: they emphasize multi-agent architecture design and production orchestration rather than on-policy RL training. For each framework, we summarize along four axes: (i) \textit{Planning \& Roles}, (ii) \textit{Tools \& Interfaces} (search/crawl/exec/MCP), (iii) \textit{Human Oversight \& Observability}, and (iv) \textit{Evaluation \& Reporting}.

These frameworks differ primarily in where planning resides (implicit recursive loop vs. explicit planner), how specialization is expressed (single agent with internal phases vs. multi-agent roles), and production affordances (e.g., MCP connectors, structured logging). These design choices drive concrete trade-offs in cost and latency, achievable depth, reproducibility, and extensibility for future tools.

\begin{table*}[ht]
  \footnotesize
  \centering
  \caption{Open-source deep research frameworks (sorted by published date). Note: GitHub stars are as observed on 06 Sept 2025.}
  \begin{tabularx}{\textwidth}{@{}p{7cm} p{2cm} >{\centering\arraybackslash}p{1.5cm} p{3cm}@{}}
    \toprule
    \textbf{Open-Source Frameworks} & \textbf{Organization} & \textbf{Stars} & \textbf{Date} \\
    \midrule
    \textit{Open Deep Research}~(\citeauthor{dzhng_deep_research_repo_2025}) & Aomni & \textasciitilde17.6k & 4 Feb 2025 \\

    \addlinespace
    \textit{DeerFlow}~(\citeauthor{bytedance_deerflow_repo_2025}) & ByteDance & \textasciitilde16.8k & 9 May 2025 \\

    \addlinespace
    \textit{Open Deep Research}~(\citeauthor{langchain_open_deep_research_repo_2025}) & LangChain & \textasciitilde8.5k & 16 Jul 2025 \\

    \addlinespace
    \textit{MiroFlow}~(\citeauthor{miromind_miroflow_repo_2025}) & MiroMind AI & \textasciitilde441 & 8 Aug 2025 \\
    \bottomrule
  \end{tabularx}
  \label{tab:deep-research-frameworks}
\end{table*}

\paragraph{System Snapshots}
\begin{itemize}
    \item \textbf{Aomni Open Deep Research.} 
    \emph{Planning \& Roles:} Single-agent recursive loop: generate queries $\rightarrow$ fetch \& extract $\rightarrow$ summarize/reflect into learnings and next directions $\rightarrow$ recurse until a depth cap, then render a Markdown report. SERP concurrency is configurable. No explicit planner/coordinator roles, i.e., planning is implicit in the next-direction proposal.\\
    \emph{Tools \& Interfaces:} Firecrawl (search + extraction); model backends via Fireworks (e.g., OpenAI, DeepSeek R1).\\
    \emph{Human Oversight \& Observability:} None; no dedicated planner review UI.\\
    \emph{Evaluation \& Reporting:} Minimal; intended as a clean baseline rather than a full product stack.

    \item \textbf{ByteDance DeerFlow.}
    \emph{Planning \& Roles:} Explicit planner decomposes tasks; a coordinator manages lifecycle; a research team of specialists (e.g., researcher, coder) executes; a reporter aggregates/formats outputs. Orchestrated with LangGraph (stateful graph).\\
    \emph{Tools \& Interfaces:} Tavily/Brave for search, Jina for crawling, optional private KB (RAGFlow), Python execution; broad MCP support.\\
    \emph{Human Oversight \& Observability:} Optional plan review/auto-accept; rich report post-editing and TTS. LangGraph Studio \& LangSmith tracing; Docker/Compose for backend + frontend.\\
    \emph{Evaluation \& Reporting:} Tests/examples provided; no standardized leaderboard.

    \item \textbf{LangChain Open Deep Research.} 
    \emph{Planning \& Roles:} A LangGraph agent with configurable subtasks (summarization, research, compression, final report). The repo also ships two \emph{legacy} topologies for comparison: (1) plan-and-execute (human-reviewable section plan) and (2) supervisor–multi-researcher (parallelism for speed).\\
    \emph{Tools \& Interfaces:} MCP-compatible; switchable across multiple model/search providers.\\
    \emph{Human Oversight \& Observability:} Supported via the plan-and-execute variant (human review of the section plan). Runs in LangGraph Studio / OAP UI.\\
    \emph{Evaluation \& Reporting:} First-class Deep Research Bench harness.

    \item \textbf{MiroMind MiroFlow.}
    \emph{Planning \& Roles:} A pipeline boots an orchestrator that manages multi-turn tool calls, delegates to sub-agents (e.g., browsing) with their own toolsets/prompts, and aligns outputs via a dedicated summarizer/formatter.\\
    \emph{Tools \& Interfaces:} Unified LLM client (OpenRouter/Anthropic/OpenAI/etc), MCP Tool Manager with FastMCP servers (search, vision, code, audio, reading, reasoning), optional E2B sandbox.\\
    \emph{Human Oversight \& Observability:} Built-in web UI for review/edit; built-in logging.\\
    \emph{Evaluation \& Reporting:} GAIA evaluation scripts with reproducible metrics (e.g., pass@1 / avg@3).
\end{itemize}

\paragraph{Design Patterns \& Trade-offs}
We synthesize differences using the same four axes to surface key trade-offs in reproducibility, throughput, and auditability.
\begin{itemize}
    \item \textbf{Planning \& Roles.} \textit{Aomni} uses \emph{implicit} planning inside a recursive loop; \textit{DeerFlow} adopts an \emph{explicit} planner with a coordinator; \textit{LangChain} supports both a single-agent graph and legacy plan-and-execute / supervisor–multi-researcher variants; \textit{MiroFlow} centralizes planning in an orchestrator with hierarchical sub-agents. This choice sets reproducibility, human-in-the-loop (HITL) insertion points, debugging granularity, and the simplicity–specialization balance (\textit{Aomni} is simple and quick to adapt, while \textit{DeerFlow}/\textit{MiroFlow} gain throughput and tool coverage from specialist teams at the cost of coordination overhead; \textit{LangChain} provides a controlled setting to compare these topologies under one roof).
    \item \textbf{Tools \& Interfaces.} \textit{DeerFlow}, \textit{LangChain}, and \textit{MiroFlow} expose broad MCP connectors; Aomni intentionally keeps a narrow interface (Firecrawl + LLM). Wider tool surfaces improve enterprise readiness but enlarge failure modes and complicate evaluation.
    \item \textbf{Human Oversight \& Observability.} \textit{DeerFlow} emphasizes plan review and report editing; \textit{DeerFlow} and \textit{LangChain} leverage LangGraph Studio; \textit{MiroFlow} ships logging and a visual UI; \textit{Aomni} is deliberately bare-bones. These choices impact auditability and developer velocity.
    \item \textbf{Evaluation \& Reporting.} \textit{LangChain} integrates \emph{Deep Research Bench}; \textit{MiroFlow} provides \emph{GAIA} scripts; \textit{DeerFlow} includes tests/examples but no standardized leaderboard; Aomni serves as a minimal baseline. Built-in harnesses enable apples-to-apples comparisons.
\end{itemize}

\paragraph{Actionable Insights}
\begin{itemize}
    \item \textbf{Use Aomni as a loop baseline.} Ideal for isolating the effects of recursion depth/breadth and for ablating query-generation and reflection strategies without multi-agent confounds.
    \item \textbf{Prototype explicit planners on DeerFlow.} The planner-coordinator-team-reporter split is a natural substrate for studying hierarchical credit assignment (planner-level vs.\ worker-level), coordinator scheduling, and HITL gating policies.
    \item \textbf{Run controlled architecture comparisons on LangChain.} Hold the environment and tool surface constant, then toggle among single-agent, plan-and-execute, and supervisor–multi-researcher graphs to quantify quality/latency/cost trade-offs.
    \item \textbf{Stress-test production affordances on MiroFlow.} The tool manager, sub-agents, and observability suite support measuring failure recovery, tool-latency masking, backpressure handling, and reproducibility under load (e.g., GAIA).
\end{itemize}

\subsection{Academic Architectures \& Coordination}
We review six representative deployment frameworks for deep research systems from academia that integrate search and other tools into agentic workflows, emphasizing multi-agent or multi-component designs and the recurring patterns behind them.

\begin{table*}[ht]
  \small
  \centering
  \caption{Representative academic systems for deep research.}
  \begin{tabularx}{\textwidth}{@{}p{3.5cm} p{3.5cm} 
  >{\raggedright\arraybackslash}X@{}}
    \toprule
    \textbf{Work} & \textbf{Domain} & \textbf{Agent Architecture} \\
    \midrule
    \textit{OWL} \citep{owl_2025} & Multi-agent assistant & Planner-Coordinator-Executors \\
    
    \addlinespace
    \textit{CoA} \citep{li2025chainofagents} & Multi-tool (web + code) & Single AFM with role-activated agents \\
    
    \addlinespace
    \textit{PaSa} \citep{he-etal-2025-pasa} & Academic search & Crawler \& Selector agents \\
    
    \addlinespace
    \textit{WebThinker} \newline\citep{webthinker_2025} & Open-web research & Single large reasoning model loop with explore/draft actions\\
    
    \addlinespace
    \textit{HiRA} \citep{hira_2025} & Deep search & Planner-Coordinator-Executors \\
    
    \addlinespace
    \textit{DeepResearcher} \newline\citep{deepresearcher_2025} & Real-world web & Single planner with a browsing helper \\
    
    \bottomrule
  \end{tabularx}
  \label{tab:acad-deep-search}
\end{table*}

\paragraph{System Snapshots}
\begin{itemize}
    \item \textbf{OWL (Optimized Workforce Learning).} Modular ``Workforce'' with a \emph{domain-agnostic Planner}, a \emph{Coordinator}, and tool-equipped \emph{Executors} (including web/search), designed to be plug-and-play across domains; only the Planner is optimized post-SFT via real-web (online DPO), boosting cross-domain generalization and achieving open-source SOTA on GAIA.

    \item \textbf{CoA (Chain-of-Agents).} Collapses a multi-agent workflow into a single Agent Foundation Model (AFM), a role-conditioned backbone that dynamically activates tool- and role-agents. Training is two-stage: (1) multi-agent distillation to produce chain-of-agents trajectories for SFT; (2) agentic RL on verifiable tasks (web + code) to refine end-to-end problem solving. Delivers strong results while avoiding joint multi-policy optimization; code, data, and model weights are open-sourced.
    
    \item \textbf{PaSa (Paper Search Agent).} Two-agent loop for scholarly discovery: a \emph{Crawler} expands a paper queue via search + citation chasing and a \emph{Selector} prioritizes candidates; training uses Imitation Learning (IL) $\rightarrow$ RL (session-level PPO within AGILE) on \emph{AutoScholarQuery}, with evaluation on \emph{RealScholarQuery}, yielding large recall gains over Google/Scholar baselines.

    \item \textbf{WebThinker.} Embeds a \emph{Web Explorer} directly into the reasoning language model's chain so the model can \emph{think-search-navigate-extract-draft} in one continuous loop; adds drafting/check/edit tools for live report writing; improves tool use via online DPO on trajectories, lifting performance on GAIA/HLE/WebWalker benchmarks.

    \item \textbf{HiRA: Decoupled Planning \& Execution.} Three-tier hierarchy, i.e., \emph{Planner} - \emph{Coordinator} - \emph{Executors}, that routes subtasks to domain-specialized executors and feeds back distilled results (not raw tool dumps); supports plug-and-play executors and dual-channel memory, improving answer quality and efficiency on complex, cross-modal deep search.

    \item \textbf{DeepResearcher.} End-to-end RL on the open web: a single \emph{planner} policy trained with GRPO learns when/how to search, browse, and synthesize over a compact JSON tool interface. Rewards are automatic (word-level F1 with format penalties), and tool outputs are treated strictly as observations (no gradient on retrieved text). A lightweight \emph{browsing helper} segments long pages and returns distilled snippets, keeping the policy focused on high-level sequencing rather than DOM-level actions. This planner+helper design—rather than a deep hierarchy like HiRA—makes end-to-end RL feasible and delivers strong real-web performance.
\end{itemize}

\paragraph{Actionable Insights}
\begin{itemize}
    \item \textbf{Hierarchical orchestration.} A planner (often via a coordinator) delegates to specialized executors and aggregates \emph{distilled} results to keep long-horizon reasoning clean and scalable~\citep{webthinker_2025,hira_2025,owl_2025,li2025chainofagents}.
  
    \item \textbf{Memory-guided iterative retrieval.} Agents refine queries across hops while maintaining scratchpads/knowledge caches and provenance, steering subsequent evidence selection and avoiding search myopia~\citep{deepresearcher_2025,he-etal-2025-pasa,webthinker_2025}.
    
    \item \textbf{Structured tool interfaces with stateful execution.} Actions are issued via compact JSON/code primitives; browsing/search helpers return structured snippets, enabling batching, retries, and reproducibility~\citep{deepresearcher_2025,webthinker_2025,li2025chainofagents}.
    
    \item \textbf{Beyond SFT: preference/RL optimization.} Targeted tuning (DPO/GRPO/PPO) of the planner or full loop improves tool use, coordination, and end quality~\citep{owl_2025,deepresearcher_2025,webthinker_2025,he-etal-2025-pasa,li2025chainofagents}.
\end{itemize}

\subsection{RL for Multi-Agent Coordination}
This section surveys RL formulations that learn two or more decision-making policies within one deep research system, optimizing from a system-level objective. Unlike the common ``single controller + fixed tools'' setup, the works here either (i) perform joint cooperative multi-agent reinforcement learning (MARL) with explicit cross-agent credit assignment, for example Multi-Agent Proximal Policy Optimization (MAPPO; \cite{yu2021mappo}) under Centralized Training with Decentralized Execution (CTDE) as in MMOA-RAG or critic-free, group-relative advantages (MHGPO), often with role-conditioned parameter sharing and decentralized execution at test time; or (ii) pursue coordinate-wise component updates via globally aligned local rewards (Optimas), which is not joint MARL but is practical when full end-to-end multi-agent training is brittle or costly. For orientation, we reference MAPPO as a standard CTDE baseline; the surveyed methods address non-stationarity, long-horizon/sparse rewards, and compute limits through centralized critics or critic-free advantages, warm-starts and constrained action spaces, conservative/trust-region updates, and shared backbones.

\begin{table*}[ht]
  \footnotesize
  \centering
  \caption{Multi-agent training strategies for deep research systems.}
  \begin{tabularx}{\textwidth}{@{}p{2.5cm} p{3.5cm} p{2.9cm} >{\raggedright\arraybackslash}X@{}}
    \toprule
    \textbf{Work} & \textbf{Trainable Roles} & \textbf{Joint end-to-end RL? (How)} & \textbf{Credit assignment / params (notes)} \\
    \midrule
    \textit{MHGPO} \newline\citep{chen2025mhgpo} &
    Query Rewriter; Reranker; Answerer &
    $\checkmark$ Critic-free (group-relative) &
    Global terminal reward; shared LLM backbone; no critic \\

    \addlinespace
    \textit{MMOA-RAG} \newline\citep{chen2025mmoa} &
    Query Rewriter; Document Selector; Answer Generator (retriever fixed) &
    $\checkmark$ MAPPO (CTDE) &
    Shared global reward; shared backbone; SFT warm-start \\

    \addlinespace
    \textit{Optimas} \newline\citep{wu2025optimas} &
    Prompts / routers / model params / hyperparams (heterogeneous) &
    $\times$ Coordinate-wise (per-component) &
    Learned LRF per component aligned to global metric; optional RL within modules \\
    \bottomrule
  \end{tabularx}
  \label{tab:rl-multi}
\end{table*}

\paragraph{MHGPO}
This paper studies a three-agent multi-hop search pipeline (Query Rewriter $\rightarrow$ Reranker $\rightarrow$ Answerer) and introduces MHGPO, a critic-free MARL method for LLM agents. It replaces value critics with group-relative advantages computed over heterogeneous rollout groupings—Independent Sampling, Fork-on-First, and Round-Robin—to balance exploration cost and signal quality. Final-answer rewards are propagated upstream; per-agent losses are then combined to update a single, role-conditioned LLM backbone, yielding joint end-to-end optimization without a critic. Empirically, it outperforms MAPPO on multi-hop QA, requires no SFT warm-up, and lowers compute/memory by removing critics. The trade-off is that all agents share one parameter set rather than maintaining distinct policy networks.

\paragraph{MMOA-RAG}
This work casts a realistic RAG pipeline as cooperative MARL and jointly trains the Query Rewriter, Document Selector, and Answer Generator with MAPPO under a shared global reward (F1/EM). It uses CTDE with a centralized critic, SFT warm starts for each role, constrained action spaces for the selector (document ID tokens), and light format penalties to stabilize exploration. All three agents share one role-conditioned LLM backbone; the retriever is fixed. Across HotpotQA, 2WikiMultihopQA, and AmbigQA, joint training of the modules outperforms single-module tuning and shows promising out-of-distribution transfer. This is bona fide multi-agent RL across the trainable modules, though parameter sharing and a fixed retriever stop short of full end-to-end optimization of the entire pipeline.

\paragraph{Optimas}
Optimas does not perform joint multi-agent RL. Instead, it learns a local reward function (LRF) for each component (prompt, router, model, hyperparameters) and adapts these LRFs online so they stay aligned with the system’s global metric; components are then improved coordinate-wise via trust-region-style updates (prompt/search/model selection, or RL when appropriate). This delivers consistent gains across compound systems while avoiding simultaneous multi-agent credit assignment. In practice, Optimas is a strong alternative when full end-to-end multi-agent RL is too brittle or costly: it can include RL within selected modules yet sidesteps the complexity of joint training.

\subsection{Discussion}
The surveyed frameworks and papers point to a converging recipe for practical deep research systems. First, separating planning from execution keeps the planner’s state clean while specialization and routing increase throughput and depth; this holds across simple single-loop designs (Aomni), explicit planner–coordinator splits (DeerFlow), configurable graphs (LangChain), and orchestrator-led hierarchies (MiroFlow). Second, structured tool interfaces and narrow, well-typed actions (search, browse, code; MCP) reduce failure modes and make recovery, caching, and retries feasible. Third, human oversight and observability are not optional: review points, trace capture, and replay make systems auditable. Finally, stronger RL-trained planners slot naturally into these stacks, improving global behavior without requiring full joint training of all components.

We highlight three open questions for this area: 
(i) Learning to coordinate. How do we train coordination itself, including credit assignment across planner, coordinator, and executors, learned communication protocols, and role discovery, while preserving stability?
(ii) Adaptive topology and scheduling. When should a system expand or contract its team, route to new specialists, or change its plan given budget, latency, and risk constraints, and can these decisions be learned rather than scripted?
(iii) Standards for portability and replay. What common schemas for actions, traces, judges, and tool results will enable reliable replay under a non-deterministic web and allow results to transfer across stacks with minimal glue?
\section{Evaluations} \label{ch:eval}
Reliable evaluation is central to measuring progress in deep research systems but remains difficult because these agents operate as multi-step, tool-using workflows with both objective and open-ended outputs. In this chapter, we map the evaluation space and focus on what current practice actually measures and how. We group the landscape into three families: question answering (QA) and vision question answering (VQA), which stress final answer accuracy under retrieval and browsing; long form synthesis, which assesses the quality of extended reports; and domain grounded agent benchmarks, which test end-to-end task execution with tools. Although evaluation is not itself an RL objective, it is essential for validating claims about training regimes, reward designs, and architectures, and for setting shared standards.

The first part of this section outlines key QA/VQA benchmarks developed to evaluate the reasoning and information retrieval capabilities of agentic systems. These benchmarks include traditional single- and multi-hop QA/VQA datasets to more advanced, LLM-driven dynamic web browsing settings, where the model must plan and adapt its search in real time. Together, they provide a comprehensive and diverse evaluation suite, enabling systematic assessment of how effectively LLM-based agents can search for relevant information, retrieve and synthesize evidence from multiple sources, reason across disparate facts, and generate accurate, contextually grounded answers for complex information-seeking scenarios.

Another key component of evaluating deep research systems involves assessing their ability to generate high-quality long-form texts, including summaries, explanations, reports, and academic-style outputs. Unlike short-form tasks with single correct answers, long-form generation is inherently open-ended and must be judged across multiple dimensions such as coherence, factuality, relevance, structure, and style. This makes evaluation particularly challenging. While human evaluation remains the gold standard, recent years have seen the development of automated metrics and domain-specific benchmarks that better capture these subjective and multi-faceted qualities. The second part of this section will survey representative benchmarks and evaluation protocols for long-form text generation.

Beyond task-specific evaluations like complex QA or long-form generation, another important strand of benchmarking targets domain-specific agents. These benchmarks aim to measure how well AI agents perform in realistic workflows, specialized domains, and applied settings, offering a more faithful assessment of their practical utility. In this section, we highlight representative benchmarks in this emerging space, which are critical for evaluating end-to-end capabilities of deep research agents under realistic operating conditions.

\begin{table*}[t]
\centering
\small
\caption{Benchmarks surveyed in this chapter and their key features.}
\begin{tabularx}{\textwidth}{@{}l >{\raggedright\arraybackslash}X@{}}
\toprule
\textbf{Benchmark} & \textbf{Key features} \\
\midrule
\multicolumn{2}{@{}l}{\textbf{\emph{QA/VQA Benchmarks}}}\\
HotpotQA & Multi-step reasoning over multiple Wikipedia documents. \\
2WikiMultiHopQA & Multi-hop questions across two Wikipedia articles. \\
Natural Questions (NQ) & Real Google queries; short and long answers. \\
MuSiQue & Multi-hop with distractor paragraphs to test robustness. \\
FEVER & Claim verification with Wikipedia evidence. \\
QASC & Science questions; multi-sentence composition across corpora. \\
Bamboogle & Search like retrieval over a static corpus. \\
FRAMES & Higher complexity with fixed corpus retrieval. \\
BrowseComp & Open web browsing; hard to find information. \\
BrowseComp-ZH & Chinese counterpart to BrowseComp. \\
InfoDeepSeek & False premise questions; detect misleading or unanswerable queries. \\
Webwalker & Multi-step gathering on real sites; query formulation and navigation. \\
WideSearch & Large-scale exhaustive gathering; breadth, consistency, fidelity. \\
MMSearch & Early multimodal search; small scale. \\
MMDocIR & Large multimodal suite (long docs, images, QA, evidence chains)  \\
MRAMG-Bench & Large multimodal suite (docs, images, QA); text and visual answers. \\
M$^{2}$RAG & Multimodal RAG: captioning, QA, fact verification, image reranking. \\
MMDocRAG & Multimodal RAG: multimodal retrieval, reranking, answer generation. \\
MM-BrowseComp & Multimodal BrowseComp; image or video signals on webpages. \\
Omni Bench & Beyond vision and language; adds audio, video, structured data. \\

\midrule
\addlinespace[0.6ex]
\multicolumn{2}{@{}l}{\textbf{\emph{Long-Form Text Benchmarks}}}\\
HelloBench & Long text queries in five categories; HelloEval (LLM judge plus checklist). \\
ProxyQA & Human-authored proxy questions capturing key points. \\
WritingBench & Query dependent criteria generated per instance. \\
LongEval & LLM as judge vs human references (arXiv, Wikipedia, blogs). \\
DeepResearch Bench & Deep research report generation; 100 PhD level tasks; RACE and FACT metrics. \\

\midrule
\addlinespace[0.6ex]
\multicolumn{2}{@{}l}{\textbf{\emph{Domain-Grounded Benchmarks}}}\\
Xbench & Real-world productivity; recruitment and marketing workflows. \\
$\tau^2$ Bench & Dual control telecom; user and agent both act with tools. \\
Finance Agent Benchmark & Financial research workflows. \\
FinGAIA & Finance benchmark in Chinese. \\
OdysseyBench & Office productivity across Word, Excel, PDF, email, calendar; long horizons and tool coordination. \\
\bottomrule
\end{tabularx}
\label{tab:benchmarks-summary}
\end{table*}

\subsection{QA and VQA Benchmarks}

\paragraph{Text-based QA benchmarks.}
Earlier multi-hop QA benchmarks are generally in the static corpus setting where retrieval is performed over a fixed, pre-defined corpus.
For example, HotpotQA~\citep{yang2018hotpotqa} and 2WikiMultiHopQA~\citep{xanh2020_2wikimultihop}
require reasoning over multiple supporting documents from Wikipedia to answer;
Natural Questions~\citep{kwiatkowski2019natural} is derived from real Google queries and requires both short and long answer extraction;
MuSiQue~\citep{trivedi2021musique} is similar to HotpotQA and 2WikiMultiHopQA, but it also tests multi-hop reasoning robustness by adding distractor paragraphs to the context. 
FEVER~\citep{thorne-etal-2018-fever} and QASC~\citep{Khot2019QASC} both emphasize fact checking and verification. The former consists of claims and evidence from Wikipedia, while the latter are from multiple text corpora.
These datasets form the foundation for evaluating the core reasoning capabilities of LLMs in a more stable, noise-controlled setting.
Later QA benchmarks began simulating a web-like environment by integrating retrieval settings that mimic search engine use. 
Representative benchmarks like Bamboogle~\citep{press2022measuring} and FRAMES~\citep{krishna-etal-2025-fact} demonstrate increased question complexity through retrieval over a static corpus to emulate real-world search behaviors. 
With the emergence of deep research systems capable of navigating dynamic and noisy web content, these QA benchmarks no longer provide sufficient challenge for meaningful evaluation.
Recent QA benchmarks have shifted toward open-web evaluation. 
BrowseComp~\citep{wei2025browsecomp} and its Chinese counterpart BrowseComp-ZH~\citep{zhou2025browsecomp} measure the ability of AI agents to locate hard-to-find information in live web environments. 
InfoDeepSeek~\citep{xi2025infodeepseek} include questions with false premise, testing whether a system can detect and handle misleading or unanswerable queries rather than producing hallucinated answers. 
Webwalker~\citep{wu2025webwalker} measures an agent’s ability to perform multi-step information gathering by traversing real websites, testing both search-query formulation and page navigation skills. 
While these benchmarks requires deep search of information, WideSeach~\citep{wong2025widesearchbenchmarkingagenticbroad} target to evaluates AI agents on large-scale, exhaustive information gathering tasks that are not requiring deep reasoning, but rather breadth, consistency and fidelity. 
Collectively, these datasets move beyond controlled, noise-free corpora to test the end-to-end skills needed for real-world, open-web information seeking, and complex problem solving. 

\paragraph{Multimodal VQA benchmarks.}
While above-mentioned benchmarks concentrate on text-only QA pairs, benchmarks that combine visual understanding with web search/browsing are rapidly maturing. 
MMSearch~\citep{jiang2025mmsearch} is one of the pioneer efforts designed to evaluate the multimodal search capability of the Large Multimodal Models (LMMs). MMSearch consists of 300 curated samples, which is really small.
MMDocIR~\citep{dong2025mmdocir} consists 1,685 questions for complex VQA on long multimodal document. It comes with not only answer for evaluating end-to-end results, but providing evidence (page and layout level) labels to evaluate intermediate deep research efficacy.
MRAMG-Bench~\citep{yu2025mramg} provides a more comprehensive multimodal benchmark. It includes 4,346 documents, 14,190 images, and 4,800 QA pairs from diverse domains, with tasks requiring both textual and visual answers. 
MMDocRAG~\citep{dong2025mmdocrag} enables evaluation on multimodal retrieval, reranking, and generation on long multimodal document. It provides 4,055 expert-annotated QA pairs with multi-page, cross-modal evidence chains, and answer in multimodal form.
M$^2$RAG~\citep{liu2025benchmarkingretrievalaugmentedgenerationmultimodal} further advances this direction by introducing a multimodal retrieval-augmented generation benchmark that spans four open-domain tasks: image captioning, multimodal question answering, fact verification, and image reranking. 
MM-BrowseComp~\citep{li2025mmbrowsecompcomprehensivebenchmarkmultimodal} is a multimodal version of BrowseComp. The questions are often prompts with images, and crucial information encountered during the search and reasoning process may also be embedded within images or videos on webpages.

Despite claiming to be multimodal benchmarks, these benchmarks involve only visual data. Omni-Bench~\citep{li2024omnibench} extends beyond vision–language tasks by integrating more modalities, including audio, video, and structured data. It is designed to assess how well agents can coordinate across heterogeneous input sources and reasoning contexts, reflecting the kinds of multi-sensory information humans process in real-world scenarios.

\subsection{Long-form Text Benchmarks}
Many user queries to deep research systems demand not just factual accuracy but well-structured, long-form text that synthesizes information across sources. Previous long-form text benchmark can serve to evaluate the text generation capabilities of the systems. 
For example, HelloBench~\citep{que2024hellobenchevaluatinglongtext} collects diverse user queries that require long text generation from different sources and divide them into five categories: open-ended QA, summarization, chat, text completion, and heuristic text generation. To evaluate the outputs, the authors developed HelloEval, a two-stage, human-aligned evaluation method, which combines LLM-as-a-judge and a checklist-based scheme which pairs each sample with 4–6 binary (yes/no) questions.
ProxyQA~\citep{tan-etal-2024-proxyqa} include manually created proxy‑questions, which are short, targeted questions probing the key points that a high‑quality answer should include, for each query. Both works rely on human annotators to create the evaluation questions. WritingBench~\citep{wu2025writingbench} takes a different approach by proposing a query-dependent evaluation framework that empowers LLMs to dynamically generate instance-specific assessment criteria. 
While these benchmarks use evaluation questions defined in advance, LongEval~\citep{alkhalifa2024longeval} uses LLM-as-a-judge to compare LLM-generated text with original human-written text from arxiv, wikipedia ang blog. 

While these benchmarks provide solid foundations for evaluating long-form text, they are not specifically designed for deep research systems. The rise of deep research systems require benchmarks that test multi-step reasoning, iterative information gathering, synthesis across diverse sources, and the ability to generate accurate, well-supported insights rather than just surface-level answers. To bridge the gap, DeepResearch Bench~\citep{futuresearch2025deepresearchbench} is the first specialized benchmark for evaluating deep research systems, with a specific focus on report generation. It covers 100 PhD-level tasks, which requires the agents to plan research steps, gather and filter evidence from diverse web sources, and synthesize it into analyst-grade, citation-rich reports. This benchmark also offers two evaluation frameworks: RACE (Reference-based Adaptive Criteria-driven Evaluation), which uses LLM-as-a-judge to score dimensions such as comprehensiveness, depth, and instruction-following against high-quality reference reports; and FACT (Factual Abundance and Citation Trustworthiness), which measures the proportion of claims backed by correct citations and the overall accuracy of these citations.

\subsection{Domain-Grounded Benchmarks}
Beyond iterative reasoning to solve complex multi-hop questions and generating high-quality long text outputs, the current deep research systems are capable of a broader range of tasks. With the integration of multiple tools, the agents are now more reliable and capable of executing domain-specific and professional-level tasks. Therefore, many researchers start to benchmarks with domain-grounded and profession-aligned evaluations, moving to more faithfully measure the utility of AI systems in practice.
Xbench~\citep{chen2025xbench} is designed to assess the real-world productivity of AI agents, with special focuse on recruitment and marketing. The evaluation tasks are shaped by professional headhunters and marketing practitioners, ensuring they reflect authentic business workflows.
$\tau^2$-Bench~\citep{barres2025tau2benchevaluatingconversationalagents} is a benchmark for assessing conversational AI in dual-control environments in Telecom domain. In the dual-control environments, both the AI agent and the user have agency and can use tools to affect a shared world. This setup is reflective of real-world scenarios such as technical support, where the user isn't merely passive but also takes actions—like toggling airplane mode or restarting a device—based on the agent’s guidance.
The Finance Agent Benchmark~\citep{bigeard2025financeagentbenchmarkbenchmarking} is a purpose-built evaluation framework designed to test LLM agents on real-world financial research challenges.
FinGAIA~\citep{zeng2025fingaiachinesebenchmarkai} also targets financial domains but it focuses on Chinese.  
OdysseyBench~\citep{wang2025odysseybenchevaluatingllmagents} targets at real-world office productivity—spanning applications like Word, Excel, PDF, Email, and Calendar. Rather than isolated tasks, it challenges agents to coordinate across multiple tools and contexts over extended time horizons.

\subsection{Discussion}
The evaluation of deep research systems has rapidly matured, moving far beyond traditional metrics of accuracy and precision. This chapter charted a clear evolutionary path for benchmarks, beginning with static, multi-hop QA datasets like HotpotQA and progressing to dynamic, open-web challenges such as BrowseComp that test real-time information seeking. The frontier has further expanded to encompass multimodal reasoning (Omni-Bench), the synthesis of high-quality long-form text (DeepResearch Bench), and most recently, performance in realistic, domain-grounded workflows (OdysseyBench, Xbench). This progression reflects a fundamental shift in the research community: from evaluating isolated skills like retrieval and reasoning to assessing holistic, system-level capabilities that mirror the complexity of real-world professional tasks. Ultimately, these sophisticated evaluation frameworks are not just measurement tools; they are crucial drivers shaping the development of more capable, reliable, and practically useful AI agents.

Though numerous benchmarks for comprehensive capability evaluation have been established as per our investigation, there still remain many potential and challenging topics to explore: 
\begin{itemize}
\item \textbf{Scalability and cost of high-fidelity evaluation.} As benchmarks become more grounded in real-world domains, cost and complexity rise sharply. Sourcing expert tasks and qualified evaluators is a major bottleneck. Future work should explore automated scenario generation and more reliable, calibrated LLM-as-a-judge systems to reduce dependence on human annotation without losing quality.
\item \textbf{Evaluating Long-Term, Interactive, and Adaptive Agents:} Most current benchmarks test discrete single-turn tasks, while real research and professional work often involve long-term projects where agents must maintain context, learn from user feedback, and adapt their strategy over multiple sessions. We need frameworks that assess these longitudinal capabilities, including memory, continual learning, and collaborative interaction.
\item \textbf{Assessing robustness, safety, and trustworthiness.} Beyond false premise checks, benchmarks should incorporate adversarial attacks, misinformation, and ethical dilemmas. Key questions include how agents handle conflicting sources, avoid harmful or biased content, and provide transparent reasoning. Standardized tests for these dimensions are essential for high stakes deployment.
\item \textbf{Beyond vision–language to true cross-modal synthesis.} Existing multimodal evaluation has focused mainly on text and images. The next step is to test reasoning across audio, video, and structured databases, with tasks that require seamless integration of heterogeneous sources to solve a single complex problem.
\end{itemize}
\section{Conclusion} \label{ch7:conclusion}
This survey focuses on RL foundations for deep research systems, covering how agents are trained end-to-end to plan, tool use, reason, and synthesize through long-horizon, tool-using interactions. Our primary scope spans three areas: data synthesis and curation for rewardable tasks, RL methods that shape decision quality over full trajectories, and systems and frameworks that make agentic RL practical and reproducible at scale. As secondary foci, we review agent architecture and coordination patterns for deployment, and evaluations and benchmarks that measure both final answers and process quality. To our knowledge, this is the first survey centered on RL for deep research, offering a unified taxonomy, aligned axes for comparing systems, and consolidated tables for quick reference. Across chapters we distill practical guidance on task construction, reward and judge design, optimizer and regime choices, and system instrumentation. We also compile discussions and open questions that matter to the community. Our aim is to give researchers and builders a compact map of the space and actionable suggestions that accelerate progress toward capable, reliable, and comparable deep research agents.

\clearpage

\bibliography{iclr2025_conference}
\bibliographystyle{iclr2025_conference}
\clearpage
\appendix

\end{document}